# A method for tissue-mask supported whole-body image registration in the UK Biobank

Yasemin Utkueri[a*], Elin Lundström[a], Håkan Ahlström[a,c], Johan Öfverstedt[a] & Joel Kullberg[a,b,c]

a Department of Surgical Sciences, Uppsala University, Sweden.
b Department of Surgical Sciences, SciLifeLab, Uppsala University, Sweden.
c Antaros Medical, Mölndal, Sweden

Corresponding author: Yasemin Utkueri

# Abstract

The UK Biobank is a large-scale study collecting whole-body MR imaging and non-imaging health data. Robust and accurate inter-subject image registration of these whole-body MR images would enable their body-wide spatial standardization, and region-/voxel-wise correlation analysis of non-imaging data with image-derived parameters (e.g., tissue volume or fat content).

We propose a sex-stratified inter-subject whole-body MR image registration approach that uses subcutaneous adipose tissue- and muscle-masks from the state-of-the-art VIBESegmentator method to augment intensity-based graph-cut registration. The proposed method (that we refer to as mask-supported) was evaluated on a subset of 4000 subjects by comparing it to an intensity-only method as well as two previously published registration methods, uniGradICON and MIRTK. The evaluation consisted of a comparison of Jacobian Determinant (JD) folding frequency, Dice scores, and voxel-wise label error frequency calculated from the 71 VIBESegmentator masks. The 40 masks from MRSegmentator and 50 masks from TotalSegmentator were also used for independent Dice score evaluations. Additionally, voxel-wise correlation between age and each of fat content and tissue volume was studied to exemplify the usefulness for medical research.

The proposed method showed 7 percentage points (pp) / 11pp lower frequency of JD folding for males / females when compared to the intensity-based method. The mask-supported method exhibited a mean Dice score of 0.773 / 0.744 across the cohort when evaluated on the 71 / 70 VIBESegmentator masks for males / females, respectively. In comparison to the intensity-only registration, the mean values were 6 pp higher for both sexes, and the label error frequency was decreased in most tissue regions. These differences were 9pp / 8pp against uniGradICON and 12pp / 13pp against MIRTK. The mask-supported method achieved a mean Dice score of 0.736 / 0.676 when evaluated with MRSegmentator and 0.683/ 0.617 when evaluated with TotalSegmentator, showing an increase that ranged between 0.7pp and 12pp from the other three methods. Using the proposed method, the age-correlation maps were less noisy and showed higher anatomical alignment.

In conclusion, the image registration method using two tissue masks improves whole-body registration of UK Biobank images.

## Introduction

Large scale epidemiological imaging studies, such as the UK Biobank [1], the German National Cohort (NAKO) [2], the Swedish CArdioPulmonary bioImage Study (SCAPIS) [3], and Uppsala Umeå Comprehensive Cancer Consortium (U-CAN) [4] collect data such as imaging, disease, and health information that can be used for disease understanding, precision medicine, or prediction tasks. While some of these studies focus on certain medical conditions, including cardiovascular and lung disease, dementia, or cancer, others aim to study health characteristics of the general population.

The UK Biobank is, to date, the largest population-wide imaging study, aiming to collect data from approximately 100,000 subjects using magnetic resonance imaging (MRI) of different anatomies, such as neck-to-knee MRI and abdominal MRI. Supporting data includes genetic information, lifestyle factors, blood samples, and disease information, which makes the UK Biobank a unique resource for epidemiological studies.

Spatial standardization of images across large cohorts allows for voxel-wise association studies as well as voxel-wise comparisons of groups of subjects [5,6]. Medical image registration is the process of finding anatomical correspondence between a reference image/space and a moving space, and is one approach for achieving such a spatial standardization. Registration can be done between scans of different subjects or scans of the same subject acquired at different times, by registering the images to a common reference space. Registration is an important component in cohort-wide voxel-wise analyses.

Deformable image registration can be performed using either iterative optimization methods or deep learning-based methods. Iterative methods search for the optimal alignment between the moving and reference images based on an objective function typically incorporating image similarity and a regularization term. One iterative method that has shown great potential is the graph-cut-based method [7, 8, 9, 10, 11], which frames image registration as a graph-cut optimization problem. Deep learning-based methods [12, 13, 14, 15], on the other hand, are trained to predict the alignment from a given image pair, optionally followed by instance-specific optimization. These models can be pre-trained for general purposes, which requires large datasets, and they can be fine-tuned for improved performance. One approach for improving image registration performance is to let segmentation masks from delineated organs and tissues guide the registration by incorporating their high-level semantic information. Organ/tissue mask-supported image registration has been shown to substantially improve performance over conventional registration methods, relying only on the image intensity information [13, 16, 17, 18].

Several general-purpose, high-performance AI-based segmentation models, such as VIBESegmentator [19], TotalSegmentator MRI [20], and MRSegmentator [21], have recently been published for application on thoracic-abdominal MR images. While some methods focus on organ segmentation, others are able to delineate specific tissues, including subcutaneous adipose tissue (SAT) and muscle. These models are able to automatically generate detailed masks of MR images without the need for manual ground truth annotation of images and training of task-specific models.

Whole-body inter-subject image registration is a challenging task due to potentially large differences in size, positioning, and topology of subjects in large cohorts, suggesting the need for robust registration methods [22]. Previous publications have shown that registration of whole-body MR images is feasible using graph-cut based [7], deep learning-based [7] and iterative b-spline [23] methods, but challenges still persist in alignment of individuals with large differences in body anatomy, particularly between abdominal organs and tissues.

Segmentation masks can be used during both the registration of medical images and evaluating the performance of the image registration. A tissue mask-supported registration method for whole-body CT images, showing satisfactory performance, has previously been introduced [16]. However, studies exploring inter-subject whole-body registration of MR images, aided by AI-generated tissue masks, have not yet been presented in the literature. Such a study, incorporating detailed evaluation of the anatomical alignment using segmentations from publicly available state-of-the-art segmentation methods, is therefore of high interest.

The aim of this study was to develop and evaluate a method that includes AI-based SAT and muscle segmentation masks to improve inter-subject registration of whole-body water-fat (Dixon) MRI. The registration method uses a graph cut-based registration method, and the masks are generated with VIBESegmentator. The evaluation comprised:

- Comparison of the proposed mask-supported method against a baseline method (without segmentation masks) to evaluate the impact of the masks on inter-subject whole-body MR image registration performance.
- Comparison of the proposed mask-supported method and baseline method against two established image registration methods (uniGradICON, MIRTK) for complementary experimental evaluation.
- Extensive and detailed evaluation of the registration quality using Jacobian Determinant (JD) folding frequency, whole-body voxel-wise statistics as well as organ and tissue-level segmentation overlap based on three independent AI-based segmentation methods (VIBESegmentator, MRSegmentator, TotalSegmentator).
- Examination of the impact of improved registration on voxel-wise correlations between age and fat content/volume.
- A sensitivity analysis of the effect of reference subject selection on registration performance.

# Materials and Methods

An overview of the proposed registration and evaluation method is given in Fig. 1.

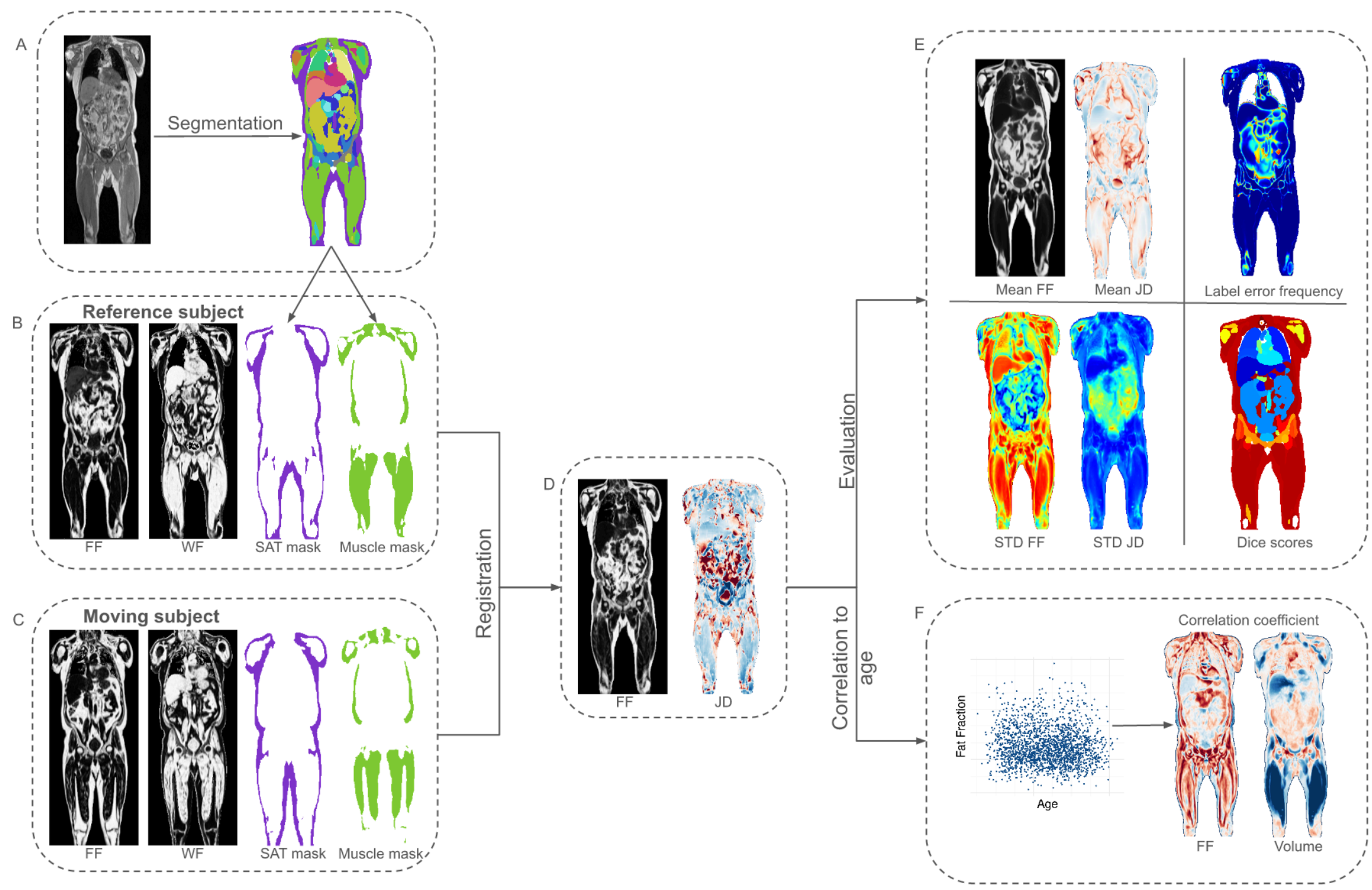

**Fig. 1**: **Illustration of the methodology and evaluation performed.** A. The summed water and fat signal images from water-fat (DIXON) MRI were segmented into 71 regions using VIBESegmentator. B. One reference subject was chosen for each sex. The fat fraction (FF) and water fraction (WF) images, and the subcutaneous adipose tissue (SAT) and muscle masks of the reference subject, chosen separately per sex, are used for every registration. C. The fat fraction and water fraction images and the SAT and muscle masks of the moving subject were used as input. D. The deformation field output from the registration was used to obtain the Jacobian determinant (JD) and the deformed fat fraction image. E. The registration methods were evaluated using different methods. 1) The mean and standard deviations (SD) of the FF and JD images were manually evaluated. 2) The Dice scores were calculated based on three independent segmentation models: VIBESegmentator (number of segmented regions = 71), MRSegmentator (40) and TotalSegmentator (50), comparing the proposed mask-supported method against three other registration methods. 3) A label error rate map was created, quantifying the error rate per-segmentation mask. F. Voxel-wise correlations of FF and volume with age were calculated and examined to show the effect of mask-supported registration in comparison to intensity-based registration. Reproduced by kind permission of UK Biobank ©

## Dataset

The dataset was composed of neck-to-knee, dual-echo DIXON image volumes from 40,296 individuals in the UK Biobank [1]. The image acquisition was conducted in 6 stations with 1.5 T MAGNETOM Aera MR-systems (Siemens Healthineers, Erlangen, Germany). The MRI protocol has been described in detail in a previous publication [24]. The UK Biobank has approval from the North West Multi-centre Research Ethics Committee with the approval number 16/NW/0274. All subjects provided informed written consent to participate at the time of inclusion. The research conducted for the present study was additionally approved by the Swedish Ethical Review Authority (DNR: 2019-03073) and performed in accordance with the Declaration of Helsinki.

The 6 stations were stitched in-house into a single whole-body volume of dimension 362 x 174 x 224 voxels. Following manual quality control, which aimed to remove images with water-fat swaps or severe artifacts (implants, missing stations, high image noise), quantitative water and fat fraction (FF) images were calculated by voxel-wise division of the water and fat signal volumes, respectively, by the sum of the water and fat signal volumes for each subject.

Initial manual quality control resulted in 1,084 volumes being filtered out, leaving 39,212 (20,310 female and 18,902 male) subjects. For this study, 2000 subjects were randomly chosen from this cohort for both males and females. These images were once again quality controlled and images with any imaging or stitching artifacts were replaced. For the ablation studies, a randomly chosen subset of 200 subjects was used for both sexes, ensuring that none of the subjects were found in the main cohort used in the study. The demographic information of the cohort and the reference subjects is given in Supplementary Table A.

## Automated AI-Based Image Segmentation

VIBESegmentator is an nnU-NET-based whole-body MR segmentation model that segments 71 tissues and organs [19, 25]. Since one of the masks is the prostate, the number of segmented regions is 70 for women. The VIBESegmentator model was trained on a subset of shoulder-to-knee volumetric interpolated images acquired during breath-hold in the NAKO and UK Biobank studies and has shown a performance of 0.92 mean Dice score across all masks and subjects on a test set. In this project, the segmentation model with ID 80 was used. The GitHub repository was cloned on the 14th of January 2025. Visual inspection of the segmentations was conducted for a small subset of subjects to ensure adequate segmentation quality. Thereafter, the summed water and fat signal images were used as input to the image registration. The SAT and muscle masks from this model were used as input for the mask-supported registration. All masks were used during the evaluation.

MRSegmentator [20] is another nnU-NET-based MR segmentation model. It segments 40 tissues and organs. Similar to VIBESegmentator, it was trained on images from UK Biobank as well as an in-house dataset and the TotalSegmentator CT dataset. The NAKO and TotalSegmenator MRI datasets were used for testing. The summed water and fat signal images were used as input for the segmentation model.

TotalSegmentator [21] is a tool for CT and MRI segmentation. It includes models trained to segment subsets of anatomical structures as well as general models. In this methodology, the total MR model was used. This model segments 50 tissues and organs. Since our images have neck-to-knee coverage, the brain segmentation was not used. Consistent with the VIBESegmentator approach, the prostate mask was excluded for female subjects, resulting in a total of 49 structures segmented for males and 48 for females.

## Image Registration

Image registration aligns multiple images to the same reference space. It allows for cohort-wise analyses, such as voxel- and supervoxel-wise regression. The Deform (v.0.5.2) [7, 8] package, is a graph-cut based deformable image registration method. It uses a multi-resolution pyramid approach by dividing the image volume into sub-regions and optimizing voxel-wise deformations in order to minimize the objective function. The optimization is done using a graph-cut approach where possible moves are evaluated for every voxel before the optimal one is chosen at every iteration. The objective function consists of the cost function, which is calculated channel-wise, a regularization term, which is used to restrict the distance at which voxels can be deformed, and a weight parameter that determines how much the regularization contributes to the objective function. In the case of multi-channel input, the sum of the objective function across all image channels is used. The method is considered to have converged when there are no possible deformations that can further decrease the objective function below a user-set threshold or after a predetermined maximum number of iterations.

One reference subject for each sex was chosen by combining a metric-based evaluation and a visual evaluation. Firstly, the BMI, weight, height, total adipose tissue (TAT), lean tissue (LT), abdominal SAT (ASAT), visceral adipose tissue (VAT), liver fat, kidney volume, liver volume, waist circumference, and hip circumference for each subject were used to calculate Z-scores based on their percentile in the cohort. Then, the 12 men and 8 women with the lowest summed absolute Z-scores were manually checked for image artifacts, body size, anatomical anomalies, and image coverage. After these two steps, one male and one female reference subject were chosen. Their images formed the fixed images, whereas the remaining subjects (moving images) were registered to these: males to the male reference image and females to the female reference image.

We propose an image registration approach that we refer to as the *mask-supported registration method,* where two binary masks, SAT and muscle, are added as input channels along with the fat fraction and water fraction (WF) channels. The SAT and muscle masks were chosen as the two input masks to the registration, as they constitute the two largest volumetric tissue compartments and provide semantic information about body composition, and the shape and position of other internal structures. After visual inspection of the 71 segmented masks, they were found to be of consistently high quality. Due to the linear increase in memory and runtime with the number of input channels in the registration, more masks were not included. Two additional channels were found to be a good balance between resource constraints and registration quality.

All four channels used channel-wise sum of squared distances (SSD) with Gaussian resampling as the cost function. The weights for the FF and WF channels were set to 1, the weights for the two mask channels to 0.6, and the regularization weight to 0.1. Six pyramid levels were used, where the final level corresponds to the full resolution of the image. The channel weights and regularization weight were chosen based on small-scale ablation studies. A table of all parameters is given in Supplementary Table A.

The output of the Deform registration method is a deformation field, which maps every voxel in the moving image to the reference space. This field is used to calculate the JD map, which is the measure of local (voxel-wise) volume change of the moving image caused by the registration. The deformation field was used to transform the FF images and masks using trilinear interpolation and nearest neighbor interpolation, respectively.

## Evaluation

Ablation studies were run to determine the optimal values for the regularization weight and the weight of the binary mask input channels. These tests were conducted on 200 randomly chosen subjects (100 men, 100 women) who were not part of the large cohort of 2000 subjects used in the rest of the analyses. For the regularization weights, 10 equally spaced values between 0.05 and 0.5 were run with the input channel weight set to 0.6. For the channel weight, 8 different values between 0.2 and 8 were tested. For both tests, the rest of the parameters were identical to the ones presented in Supplementary Table B. The best performing regularization weight in the first ablation test was used for the channel weight tests. The Dice score and Hausdorff distances were used as evaluation metrics as well as mean images for visual confirmation of the results. For the mask channel weight parameter, the mean number of voxels outside of the reference body mask was used as an additional metric after an artifact was spotted in the mean FF images. This percentage was calculated by counting the voxels in every deformed FF image outside of the body mask of the reference subject.

To evaluate the proposed registration method, its performance was compared to that obtained with the same registration method but without the support of binary masks. This latter, non-mask supported method is referred to as the *intensity-based method*.

The runtimes of the mask-supported and intensity-based methods were measured starting from the loading of the images until the deformation of the FF image for each pair of reference and moving subjects. For uniGradICON, the measurement included the registration and warping of the FF image. For MIRTK, we measured the registration of one pair of moving and reference images, including both the affine and deformable registration steps. The runtime for one pair of reference and moving images was 3 minutes on average for the mask-supported method, 2.5 minutes for the intensity-based method, 97 seconds for uniGradICON, and 84 seconds for MIRTK. All runtimes were calculated on an Intel Core i9-13900K machine with 64 GB of RAM, and an NVIDIA RTX A4000 GPU. The GPU was only used for uniGradICON and MIRTK. The mask-supported method uses 2 GB of RAM per image pair.

In the first evaluation step, the deformed FF images were visually inspected. The mean and standard deviation of both the registered FF images and the JDs were calculated for both sexes.

The JDs were evaluated by calculating the rate of folds ( defined as JD < 0 ) within the body mask of the reference subjects. This percentage was calculated across the 2000 subjects for each sex. A lower number of folds indicates a more successful registration.

To evaluate the registration performance at the tissue and organ level, the Dice scores between each subject and the reference subject were calculated for the 70-71 structures segmented using VIBESegmentator, the 40 structures from MRSegmentator, and the 49-48 structures from TotalSegmentator. The VIBESegmentator structures include the two masks that were used during the registration (SAT and muscle). During analysis, these two masks were kept separate in order to avoid bias.

The deformation fields generated in the registration process were used to move the segmentation masks into the reference space using nearest-neighbor interpolation, and the segmentations of the fixed image were used as the ground truth. Then, the Dice scores were calculated for each subject and mask.

Thereafter, to quantify and visualize the location of mask-overlap errors, we introduce a new concept called *label error frequency maps (LEFM),* which aims to depict the voxel-wise percentage of mismatched labels, i.e., the percentage of subjects in the cohort that, for a given voxel, have a different label than the reference label. These maps were computed by voxel-wise comparison of the moved segmentation labels of each subject to those of the reference subject. The segmentations from VIBESegmentator were used for this analysis because they contain more structures, including larger tissues like SAT, VAT, and muscle.

Two other previously published image registration methods of particular relevance to our work were chosen to compare against, for complementary experimental evaluation. The first was uniGradICON [12], a state-of-the-art deep learning-based foundation model for medical image registration, trained on different modalities (MRI and computed tomography) and with demonstrated high performance on internal and external test sets. In our study, the summed water-fat signal images were given as input to this model, a decision based on initial tests on a small number of subjects using different images as input, and evaluated by visual inspection. The second method was proposed by Starck et. al. in a recently published paper [23], which also used whole-body MR images from the UK Biobank cohort, and for which the authors optimized the registration parameters. The present study reused these optimized registration parameters. The Starck method proposes a population-specific atlas creation pipeline using deepali [26], a GPU-accelerated implementation of an established registration method, MIRTK [11]. MIRTK is an iterative approach to registration where an affine transformation is followed by a B-spline-based deformable transformation. Starck et al. have shown that this method has higher performance in terms of Dice score for liver, spleen, and kidneys, in comparison to VoxelMorph [13]. Following the methodology proposed in Starck et al.'s paper, the water images were used as the input for this method. The Dice scores from all three segmentation methods were used to compare our proposed method against the two previously published methods.

A correlation study of FF and local volume (represented by the JD) with subject age at time of imaging was conducted using registered images from both the mask-supported and the intensity-based registration methods to evaluate the effect of improved registrations on the correlations. The age was obtained by taking the difference between the date of birth and the date of imaging provided by the UK Biobank in month resolution. The intensity-based and mask-supported registrations were compared. Pearson correlation coefficients and adjusted p-values were calculated separately for the female and male cohorts. For every voxel, only the subjects with a non-zero voxel intensity in the FF image were included in the correlation analyses. To quantify the difference in age-correlation performance between the two methods, the percentage of voxels with statistically significant correlation within the reference subject's body mask was calculated.

## Sensitivity Analysis

The effect of choosing different reference subjects was analysed by small-scale registration tests on four alternative reference subjects for each sex. The 200 subjects from the ablation study subset were registered using the mask-supported method, intensity-based method, uniGradICON, and MIRTK. Dice scores were used as the main metric for evaluating the quality of the registrations. For both sexes, reference subjects were selected to represent different BMI groups. One subject was chosen from the 25$^{th}$ quantile, one from the 75$^{th}$ quantile, and two near the median BMI value, which are called the "medium BMI" subjects.

## Statistical testing

For the Dice score comparisons, paired Wilcoxon Signed-Rank tests with Bonferroni correction were used for statistical testing. The data fits the assumptions of this test as it does not assume normality. Additionally, in each test, we are comparing two Dice scores for each subject using different registration methods and want to test for significant differences in performance. For the age correlations, the Pearson correlation coefficient was calculated for each voxel. False Discovery Rate (FDR) correction was used as a multiple testing correction using the Benjamini-Hochberg procedure.

# Results

## Ablation studies

The ablation studies showed that a regularization weight of 0.1 was optimal in terms of both the Dice score and Hausdorff distances for both the male and female cohorts (Supplementary Figure A). For the weight of the binary mask channels, the best value in terms of Dice score and Hausdorff distance was 2 (Supplementary Figure B). After manual inspection of the mean image of the deformed FF images, however, artifacts were found. The voxel-wise mean values were calculated by correcting for the number of subjects that had a non-zero value for the specific voxel. This correction highlighted artifacts occurring in only a small number of subjects, which would otherwise have remained undetected. Supplementary Figure C shows these artifacts for the male and female cohorts. Supplementary Figure D shows the mean percentage of voxels in the FF images outside of the reference subject's body mask. Upon inspection of these mean images, and taking into account both the percentage of voxels outside of the body mask and the Dice scores for different mask channel weights, 0.6 was chosen instead of 2, as it maintained relatively high Dice scores and low Hausdorff distances while removing the observed artifacts.

## Evaluation in comparison to the intensity-based method

### Evaluation of mean and standard deviation images

Figure 2 shows the mean FF and JD images for both sexes from intensity-based and mask-supported registrations, as well as subtraction images (mask-supported - intensity-based) from these two registration methods. In the mean FF images, the largest differences between the registration methods were observed in the arms and the borders between different tissues. In the mean JD image, the largest differences were observed in the same regions as for the mean FF images, as well as in the near surroundings of certain internal organs.

As seen in Figure 3, the standard deviation images of FF and JD for both sexes showed relatively large differences related to the image registration method, in both the arms and the region close to the urinary bladder. For females, the standard deviation of FF and JD in visceral adipose tissue (VAT) was high for both registration methods, while for males, corresponding standard deviations reached their highest value lateral to the left kidney.

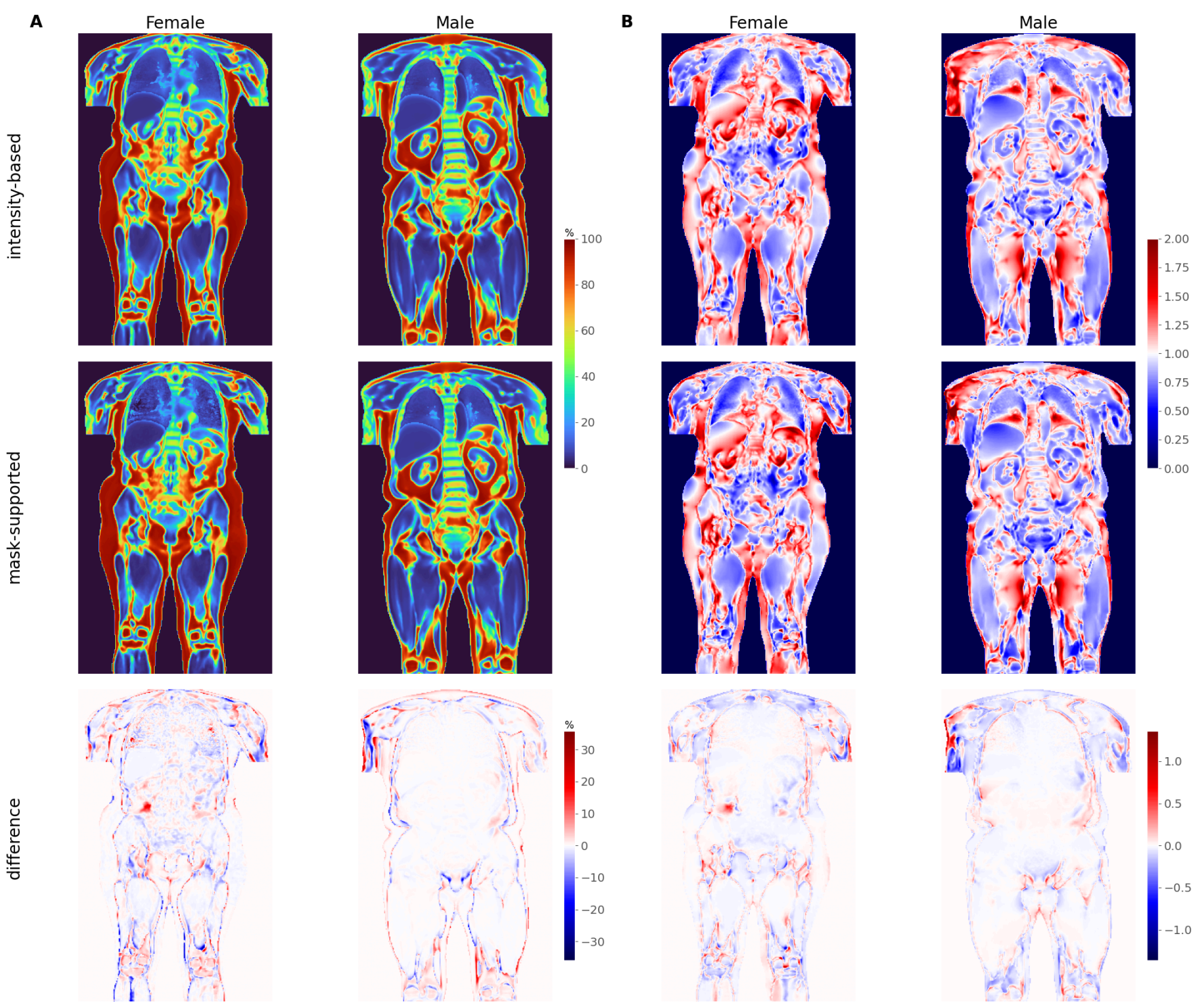

**Fig 2:** Mean FF (A) and JD (B) images for the two sexes from intensity-based and mask-supported registrations, as well as subtraction images (mask-supported - intensity-based) from the two registration methods, illustrating their differences. Reproduced by kind permission of UK Biobank ©.

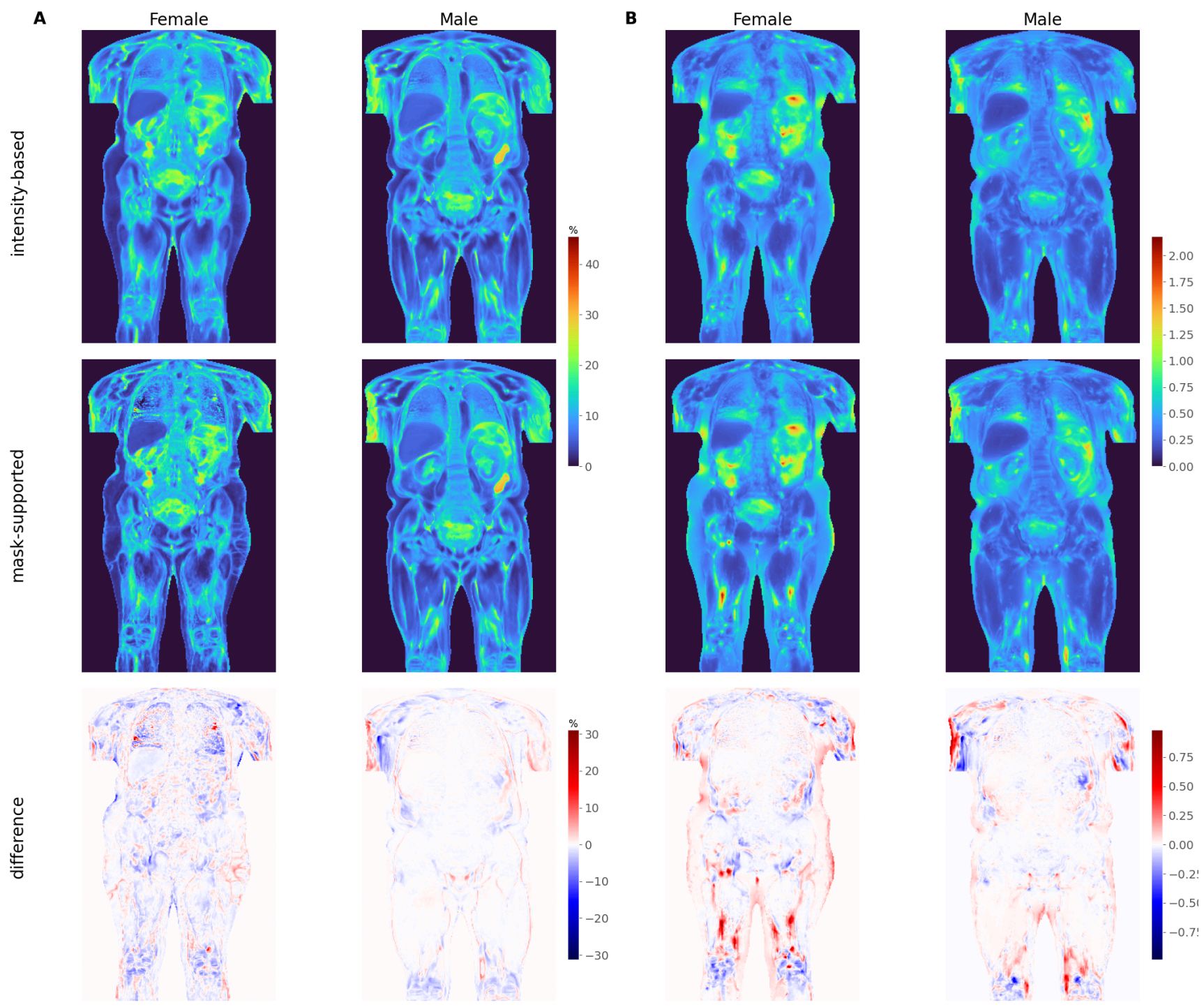

**Fig 3:** Standard deviation FF (A) and JD (B) images for the two sexes from intensity-based and mask-supported registrations, as well as subtraction images (mask-supported - intensity-based) from the two registration methods, illustrating their differences. Reproduced by kind permission of UK Biobank ©.

### Evaluation of folds in the Jacobian Determinant

The percentage of folds in the JDs for the intensity-based registrations was 0.53% and 0.51% for males and females, respectively. These numbers decreased to 0.46% and 0.40% with the mask-supported registration. Per-voxel folding rate maps are provided in Supplementary Figure E. For both sexes, the folding rate decreases in the abdominal muscles, knees, and VAT with the mask-supported registration.

### Evaluation of Dice scores

A comparison between the intensity-based and mask-supported registration methods, evaluated using the Dice scores from the VIBESegmentator organ/tissue masks, is illustrated by the spider plots in Figures 4 and 5. In these figures, the masks were grouped according to tissue or organ type based on anatomical location for improved visualization and interpretability. Supplementary Table C presents the mean Dice scores and potentially significant differences in Dice score between intensity-based and mask-supported registrations. The mask-supported registration showed significantly higher Dice scores for 63 and 60 segmentation masks (for males and females, respectively) compared to the intensity-based registration (adjusted p-value < 0.0001). The mean Dice score (excluding the two masks used as input in the registration: SAT and muscle) was 0.773 (95% CI = 0.772 - 0.774) in males and 0.744 (0.743 - 0.745) in females for the mask-supported method, whereas 0.715 (0.714 - 0.716) in males and 0.693 (0.693 - 0.695) in females for the intensity-based method. For males, the largest difference in Dice score was obtained for the thyroid gland (difference in Dice = 34 percentage points (pp)), clavicula right (33 pp), and common carotid artery right (32 pp) masks. For females, the largest difference was observed for the thyroid gland (25 pp), scapula left (22 pp), and clavicula left (21 pp) masks. The mean Dice score across all masks (including SAT and muscle) was 0.778 (0.778 - 0.779) in males and 0.75 (0.75 - 0.752) in females for the mask-supported method, whereas 0.719 (0.719 - 0.721) in males and 0.699 (0.698 - 0.701) in females for the intensity-based method. The SAT and muscle masks both showed significant improvement in males and females using the mask-supported method in comparison to the intensity-based method.

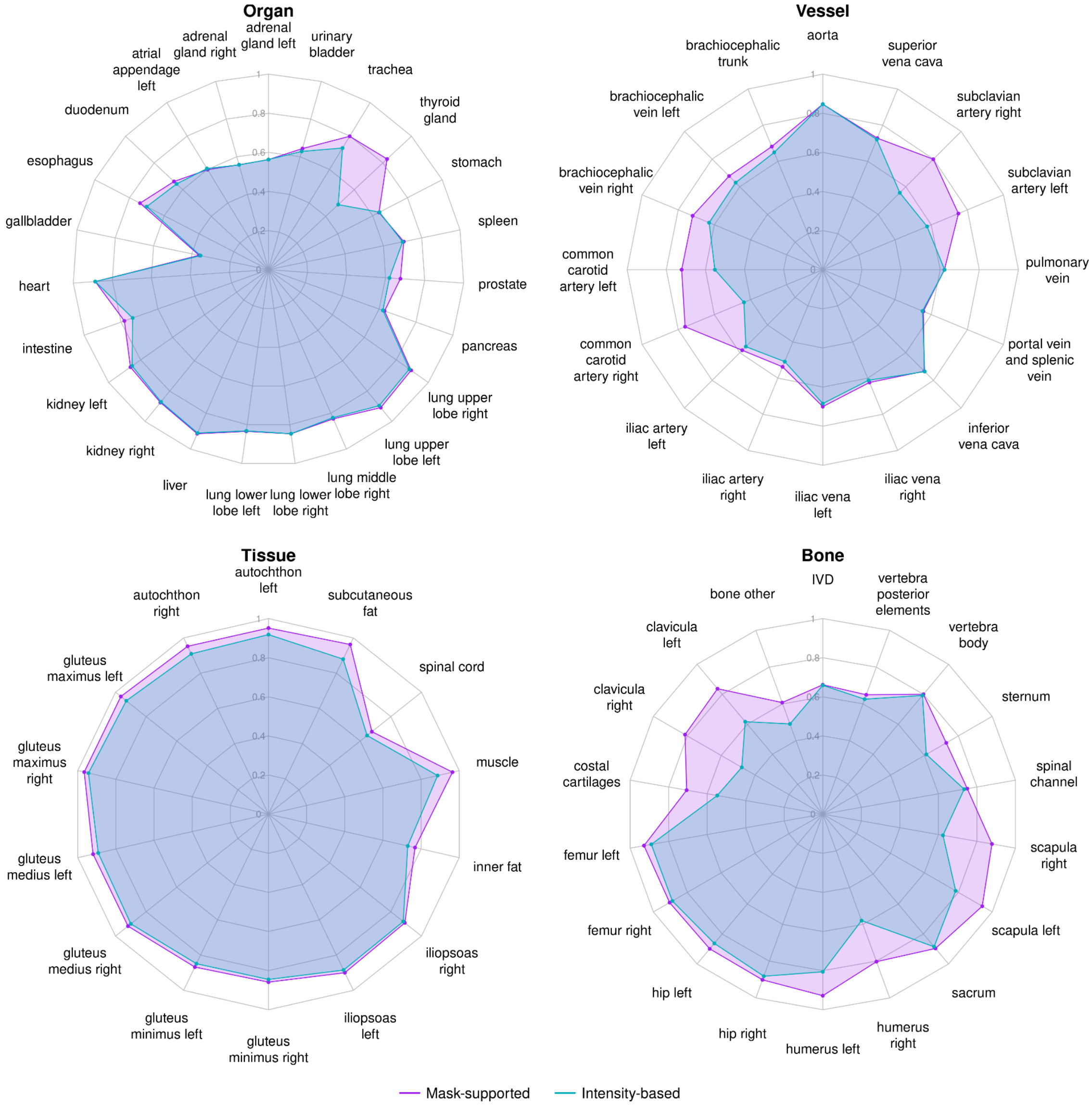

**Fig 4:** Mean Dice score comparisons between the intensity-based and mask-supported registrations as illustrated by radar plots for the 71 masks in the male cohort. The masks were separated into four groups based on anatomical classification. Dice scores are represented by the distance from the center of the plot.

**Fig 5:** Mean Dice score comparisons between the intensity-based and mask-supported registrations as illustrated by radar plots for the 70 masks in the female cohort. The masks were separated into four groups based on anatomical classification. Dice scores are represented by the distance from the center of the plot.

When evaluated with MRSegmentator, the intensity-based registration had a mean Dice score of 0.729 (0.727 - 0.73) / 0.666 (0.664 - 0.668) for males / females, respectively, while the mask-supported registration had mean Dice scores of 0.736 (0.735 - 0.737) / 0.676 (0.674 - 0.678). The mask-supported registration showed significantly higher mean Dice scores for 34 masks in males and 32 masks in females. Supplementary Table D shows the per-mask mean Dice scores and adjusted p-values.

When evaluated with TotalSegmentator, the intensity-based registration had a mean Dice score of 0.660 (0.659 - 0.662) / 0.597 (0.595 - 0.599) for males / females, respectively, while the mask-supported registration had mean Dice scores of 0.683 (0.682 - 0.685) / 0.617 (0.615 - 0.619). The mask-supported registration showed significantly higher mean Dice scores for 40 masks in males and 39 masks in females. Supplementary Table E shows the per-mask mean Dice scores and adjusted p-values.

## Evaluation of percentage label error frequency maps

Figure 6 shows the percentage label error frequency maps for both sexes. The dotted lines in the coronal slice indicate the position of the axial slices. For males, the highest error rate was seen in the urinary bladder and below the left lung. For females, it was located in the VAT and the intestines. For both sexes, most regions showed a lower label error rate in the mask-supported registration compared to the intensity-based registration. The largest benefits of the mask-supported model compared to the intensity-based model were observed in boundaries between the SAT and muscle masks, as well as in the shoulders, clavicles, and neck regions. These regions also showed the largest benefits in terms of Dice score for both sexes.

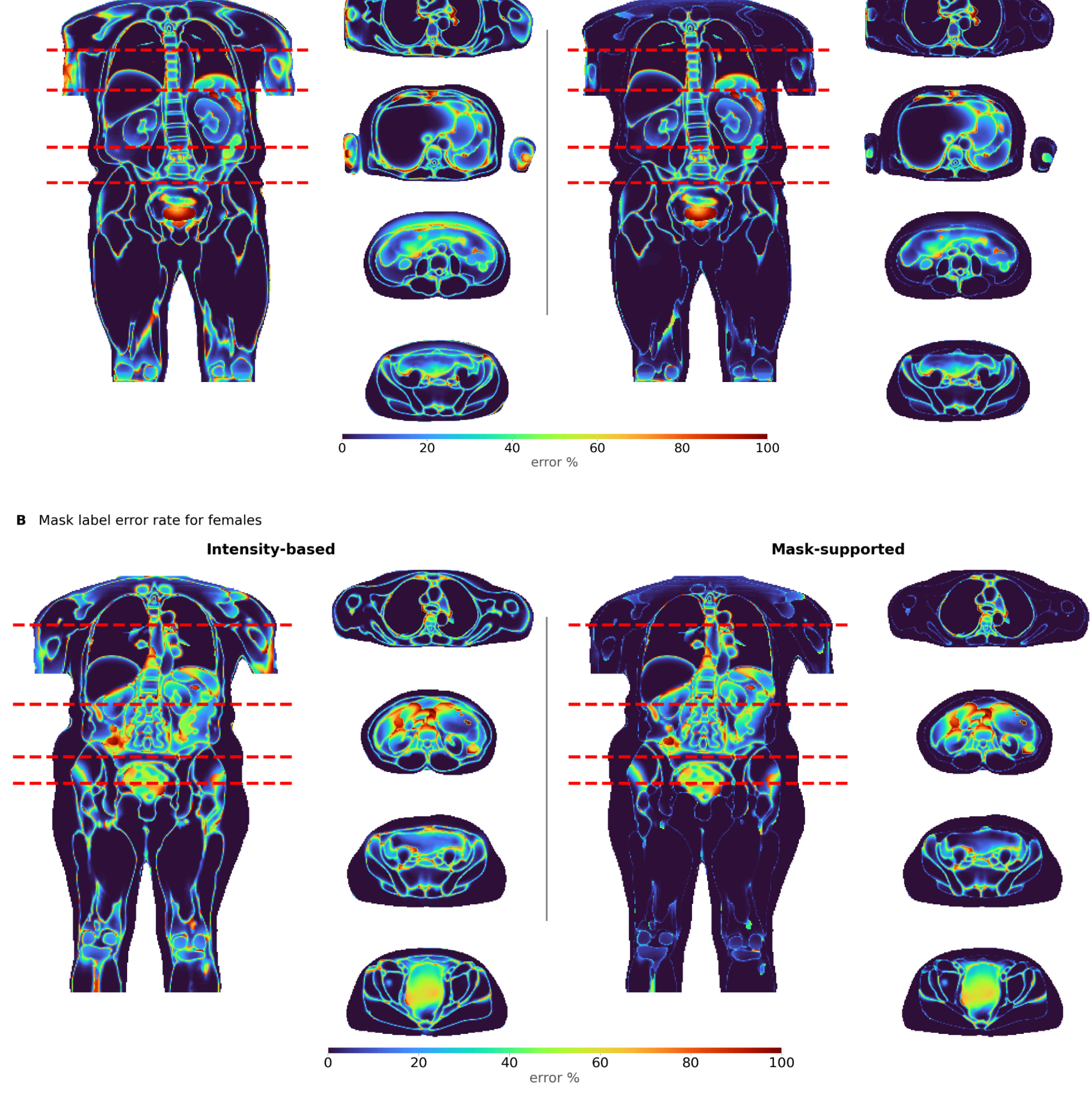

**Fig 6:** Percentage label error frequency map for males (A) and females (B) in the intensity-based and mask-supported registrations. Reproduced by kind permission of UK Biobank ©.

## Evaluation in comparison to uniGradICON and MIRTK

A comparison between the organ/tissue mask Dice scores obtained using the VIBESegmentator segmentations for the mask-supported registration and two previously published methods, uniGradICON and MIRTK, is presented in Supplementary Tables G and K, respectively. Figures 7 and 8 show the same comparisons as spider plots separately for males and females, with the segmented masks split into anatomical groups. The mean Dice scores across all 69 / 68 masks (without including SAT and muscle) for the males/females were 0.677 (0.676 - 0.679) / 0.664 (0.663 - 0.666) for uniGradICON and 0.653 (0.653 - 0.655) / 0.604 (0.604 - 0.606) for MIRTK. The mask-supported method improved the mean Dice scores by 9 pp / 8 pp compared to uniGradICON and 12 pp / 13 pp compared to MIRTK. The mask-supported method showed statistically significant improvement for 62 and 63 masks for males and females, respectively, in comparison to uniGradICON, while the numbers were 69 and 68 in comparison to MIRTK. The tissue segmentations showed consistently higher performance in the mask-based registration. The combined portal vein and splenic vein mask in females was the only mask showing significantly worse performance in the mask-supported registration compared to both uniGradICON and MIRTK. The mean Dice scores across all masks (including SAT and muscle) for the males/females were 0.683 (0.682 - 0.684) / 0.671 (0.67 - 0.672) for uniGradICON and 0.657 (0.657 - 0.659) / 0.611 (0.61 - 0.612) for MIRTK.

When evaluated with MRSegmentator, uniGradICON had a mean Dice score of 0.694 (0.692 - 0.696) / 0.631 (0.629 - 0.633) for males / females, respectively, while the mask-supported registration had mean Dice scores of 0.736 / 0.676. The mask-supported registration showed significantly higher mean Dice scores for 35 masks in males and 31 masks in females. MIRTK had mean Dice scores of 0.616 (0.614 - 0.618) / 0.561 (0.559 - 0.563), with 40 / 39 masks being significantly higher with the mask-supported method. Supplementary Tables H and L show the per-mask mean Dice scores and adjusted p-values for the comparisons between the mask-supported method and uniGradICON and MIRTK, respectively.

When evaluated with TotalSegmentator, uniGradICON had a mean Dice score of 0.630 (0.628 - 0.631) / 0.565 (0.563 - 0.567) for males / females, respectively, while the mask-supported registration had mean Dice scores of 0.683 / 0.617. The mask-supported registration showed significantly higher mean Dice scores for 42 masks in males and 37 masks in females. MIRTK had mean Dice scores of 0.575 (0.573 - 0.577) / 0.502 (0.5 - 0.503), with 48 / 47 masks being significantly higher with the mask-supported method. Supplementary Tables I and M show the per-mask mean Dice scores and adjusted p-values for the comparisons between the mask-supported method and uniGradICON and MIRTK, respectively.

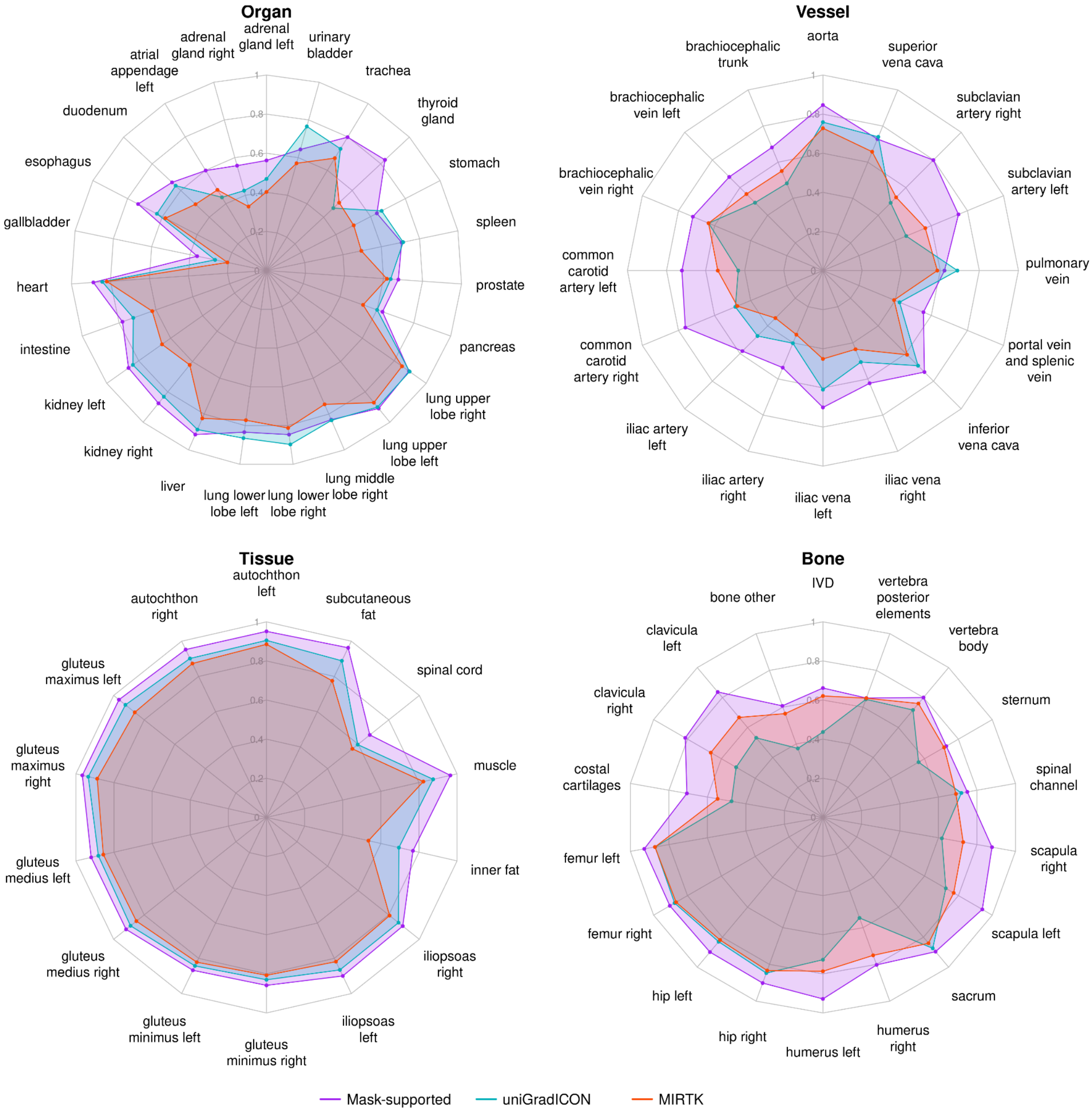

**Fig 7:** Mean Dice score comparisons for the 71 masks for the male cohort between the mask-supported, uniGradICON, and MIRTK registrations. The masks were separated into four categories based on anatomical classification.

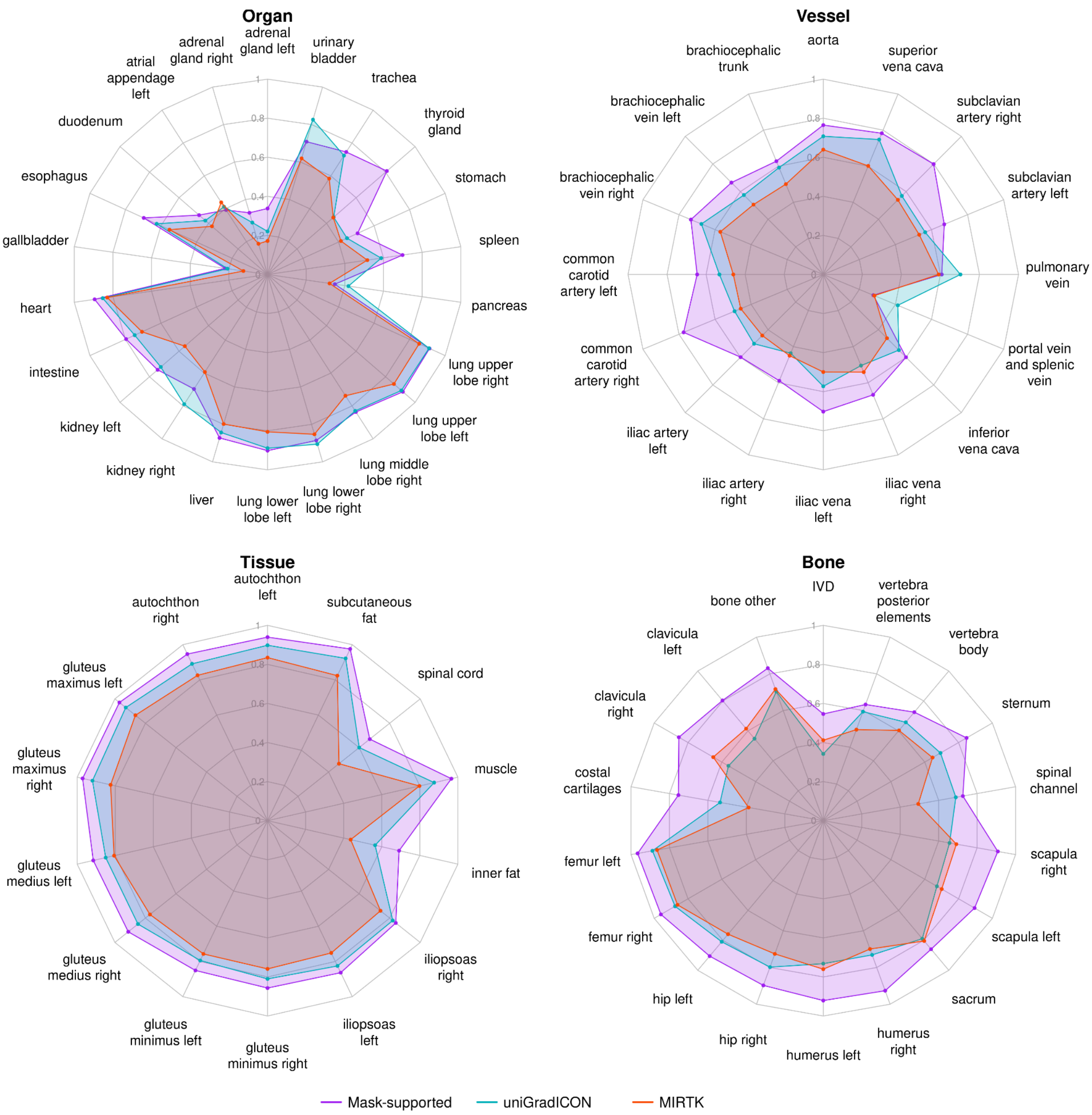

**Fig 8:** Mean Dice score comparisons for the 70 masks for the female cohort between the mask-supported, uniGradICON, and MIRTK registrations. The masks were separated into four categories based on anatomical classification.

## Reference subject sensitivity analysis

Table 1 shows the mean Dice scores for the four reference subjects that were tested, as well as the main reference subject used for the rest of the analysis presented. Scores for all three segmentation methods were presented along with the standard deviation across subjects. The BMI for every reference subject is also given.

In males, the lowest Dice scores were obtained for the high-BMI reference across all three segmentation methods (VIBESegmentator: 0.7538; MRSegmentator: 0.6991; TotalSegmentator: 0.6551). The highest mean Dice score was obtained for the main reference for VIBESegmentator and MRSegmentator (0.7878, 0.7306) and one of the medium BMI subjects for TotalSegmentator (0.6790).

For females, the lowest mean Dice score was attained for the low BMI reference using VIBESegmentator and MRSegmentator (0.7347, 0.6528) and one of the medium BMI references for TotalSegmentator (0.5991). The highest mean score achieved was for one of the medium BMI references for VIBESegmentator and MRSegmentator (0.7770, 0.7061), and one of the medium BMI references for TotalSegmentator (0.6274).

The standard deviations ranged between 0.0312 and 0.0499 for VIBESegmentator, 0.0369 and 0.0727 for MRSegmentator, and 0.0311 and 0.0648 for TotalSegmentator.

| Sex | Reference | BMI | VIBESegmentator | MRSegmentator | TotalSegmentator |
|---|---|---|---|---|---|
| **Male** | main reference | 27.9 | 0.7878 ( 0.0397 ) | 0.7306 ( 0.0727 ) | 0.6783 ( 0.0648 ) |
| | low BMI | 24.5 | 0.7720 ( 0.0383 ) | 0.7134 ( 0.0549 ) | 0.6574 ( 0.0430 ) |
| | medium BMI | 26.7 | 0.7846 ( 0.0390 ) | 0.7208 ( 0.0626 ) | 0.6780 ( 0.0439 ) |
| | medium BMI | 26.7 | 0.7831 ( 0.0409 ) | 0.7246 ( 0.0608 ) | 0.6790 ( 0.0506 ) |
| | high BMI | 29.2 | 0.7538 ( 0.0364 ) | 0.6991 ( 0.0622 ) | 0.6551 ( 0.0361 ) |
| **Female** | main reference | 22.7 | 0.7499 ( 0.0351 ) | 0.6789 ( 0.0369 ) | 0.6209 ( 0.0343 ) |
| | low BMI | 22.8 | 0.7347 ( 0.0414 ) | 0.6528 ( 0.0495 ) | 0.6022 ( 0.0381 ) |
| | medium BMI | 25.1 | 0.7523 ( 0.0358 ) | 0.6836 ( 0.0427 ) | 0.5991 ( 0.0332 ) |
| | medium BMI | 26.7 | 0.7770 ( 0.0312 ) | 0.7061 ( 0.0414 ) | 0.6274 ( 0.0311 ) |
| | high BMI | 28.3 | 0.7467 ( 0.0499 ) | 0.6679 ( 0.0667 ) | 0.6173 ( 0.0566 ) |

**Tab 1:** Mean Dice scores for 200 subjects registered to different reference subjects with the mask-supported method. Mean Dice scores for VIBESegmentator, MRSegmentator, and TotalSegmentator segmentations are presented with the standard deviation across subjects given in parentheses. The BMI of every reference subject is also shown. The main reference is the reference subject used in the rest of the analyses.

## Voxel-wise age correlation analysis

Figures 9 and 10 show correlation maps based on the voxel-wise Pearson correlation coefficients between age and each of FF and JD, separately for males and females. The correlation maps were analyzed for both intensity-based and mask-supported registrations, and only the statistically significant correlation coefficients are shown for the chosen coronal and axial slices. The dotted lines in the coronal slice indicate the position of the axial slices. Regions showing differences between the two methods are denoted by numbered arrows.

In addition to comparisons with the mask-supported method, the intensity-based method was also evaluated against uniGradICON and MIRTK. These results are shown in Supplementary Tables F and J for uniGradICON and MIRTK, respectively. After registration with the intensity-based method, 54 / 49 masks showed significantly higher Dice scores for males / females, respectively, when compared to uniGradICON. The number of significantly improved masks was 58 / 62 for males / females when compared to MIRTK.

In the FF correlation maps for both males and females, abdominal muscles align more closely with the reference anatomy in the mask-supported registration compared to the intensity-based registration (arrow 1). In addition, in females, some edges and regions of the muscle tissue (arrow 2), as well as the abdominal SAT (arrow 3), did not show significant correlation coefficients in the intensity-based registration as opposed to the mask-supported registration. For both registration methods, positive correlation with FF was seen in the muscles, whereas negative correlation was observed in adipose tissue, most significantly in abdominal SAT.

The JD correlations showed larger differences between the two registration methods compared to the FF correlations. In males, a positive correlation between VAT and age was observed in more voxels in the mask-supported method (Arrow 1). The mask-supported registration also showed a significant negative correlation between JD and age in abdominal SAT (arrow 2) and muscle (arrow 3), while the intensity-based registration showed a non-significant correlation in these regions. In females, the intensity-based registration did not show any significant correlation with age in abdominal SAT, while the mask-supported registration showed a negative relationship (arrow 1).

The comparison of the percentage of voxels within the body mask that showed statistically significant correlations for the intensity-based and mask-supported registrations are shown in Table 2. All four pairs of comparisons showed an increase in percentage with the mask-supported method.

| | **male** | | **female** | |
|---|---|---|---|---|
| | **FF** | **JD** | **FF** | **JD** |
| **intensity-based** | 46.46 | 52.54 | 44.42 | 52.98 |
| **mask-supported** | 48.66 | 53.78 | 61.52 | 54.47 |

**Tab 2:** Percentage of voxels with statistically significant correlations for the intensity-based and mask-supported registrations for the two sexes and both FF and JD.

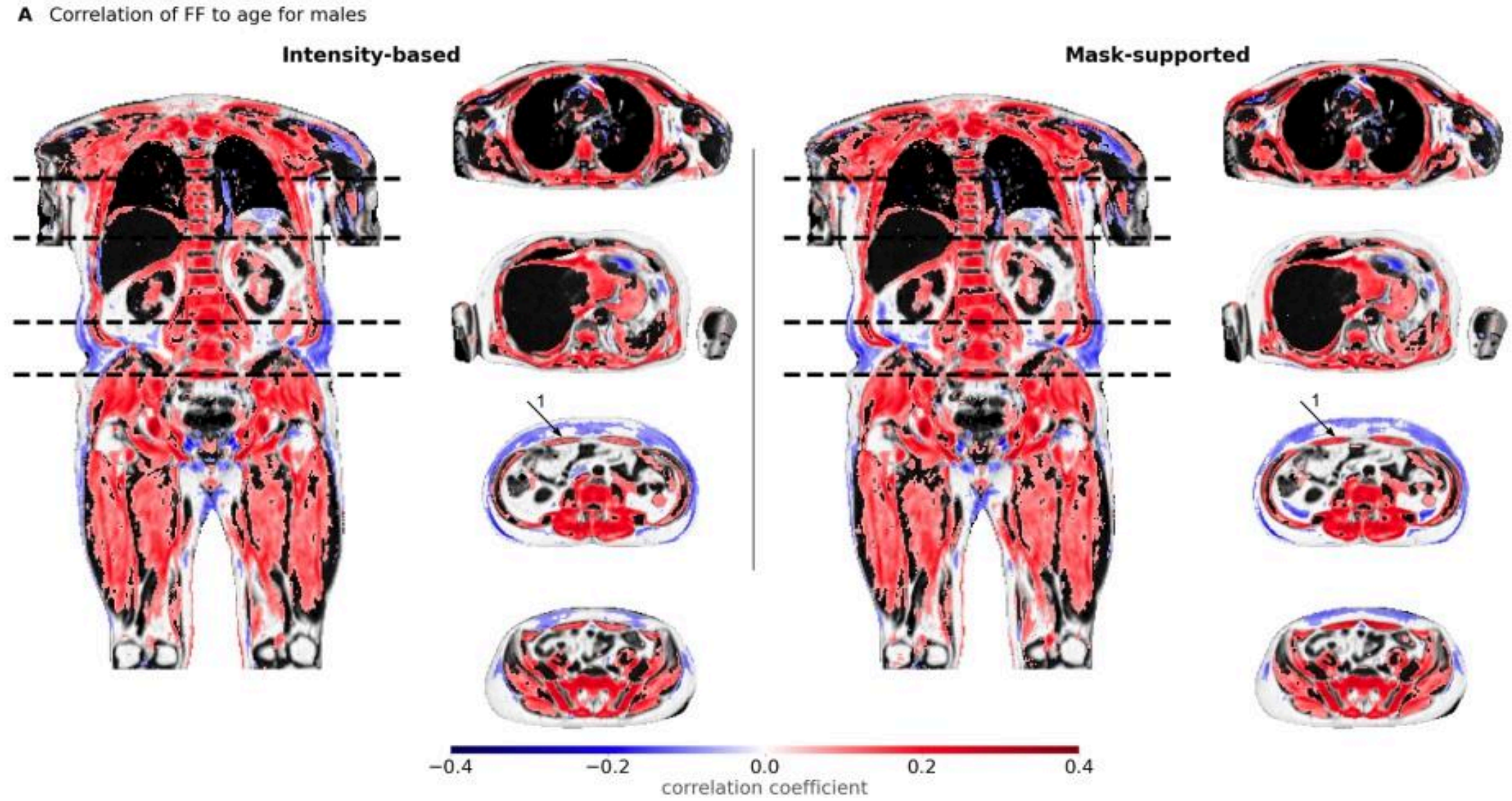

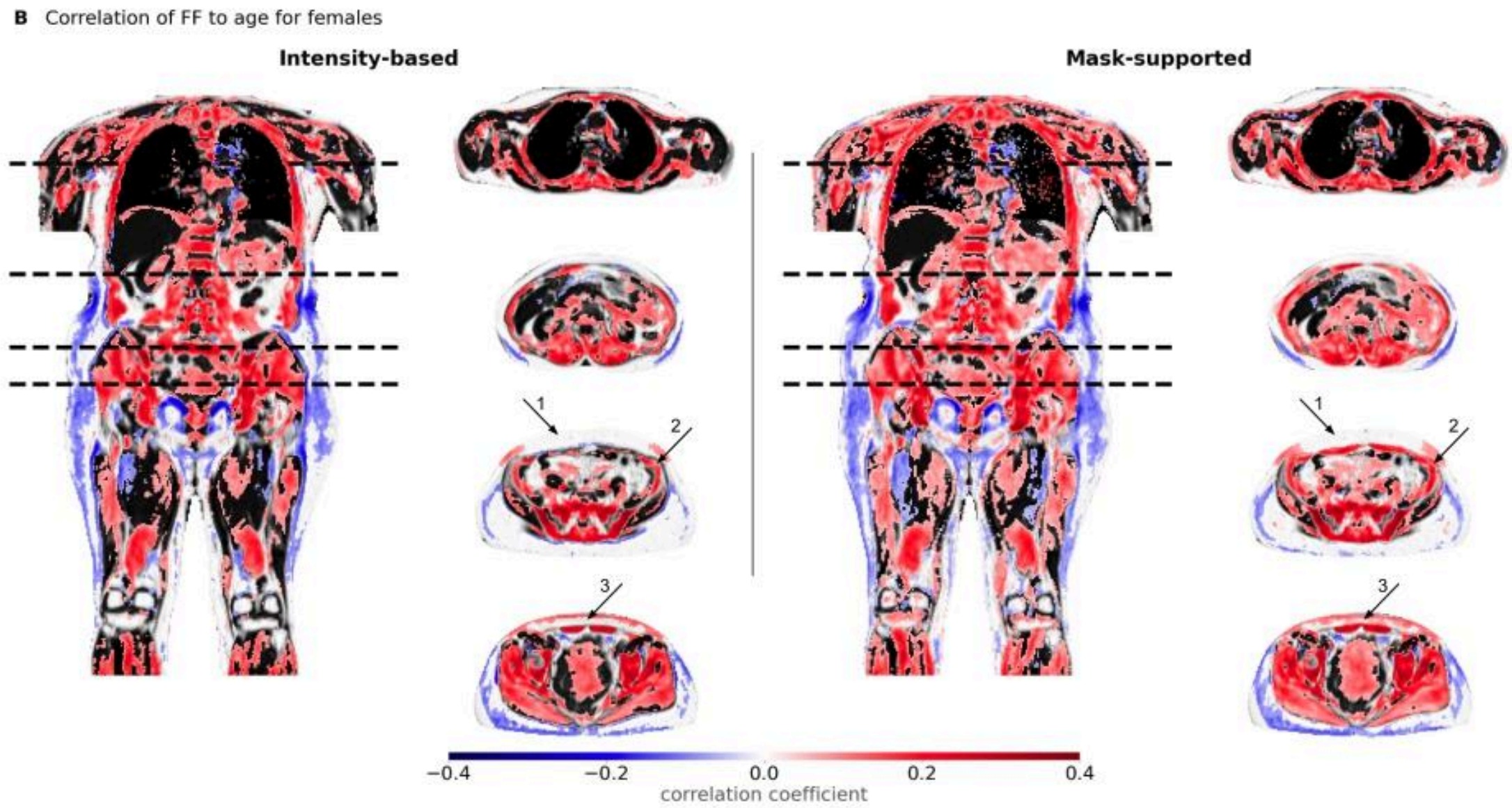

**Fig 9:** Correlation maps showing the voxel-wise Pearson correlation coefficient between age and FF for the intensity-based and mask-supported registrations, for the males (A) and females (B). Only significant correlation coefficients based on FDR-adjusted p-values are displayed. The correlation maps are overlaid on the reference subject FF images. The dotted horizontal lines in the coronal image indicate the position of the axial images. Reproduced by kind permission of UK Biobank ©.

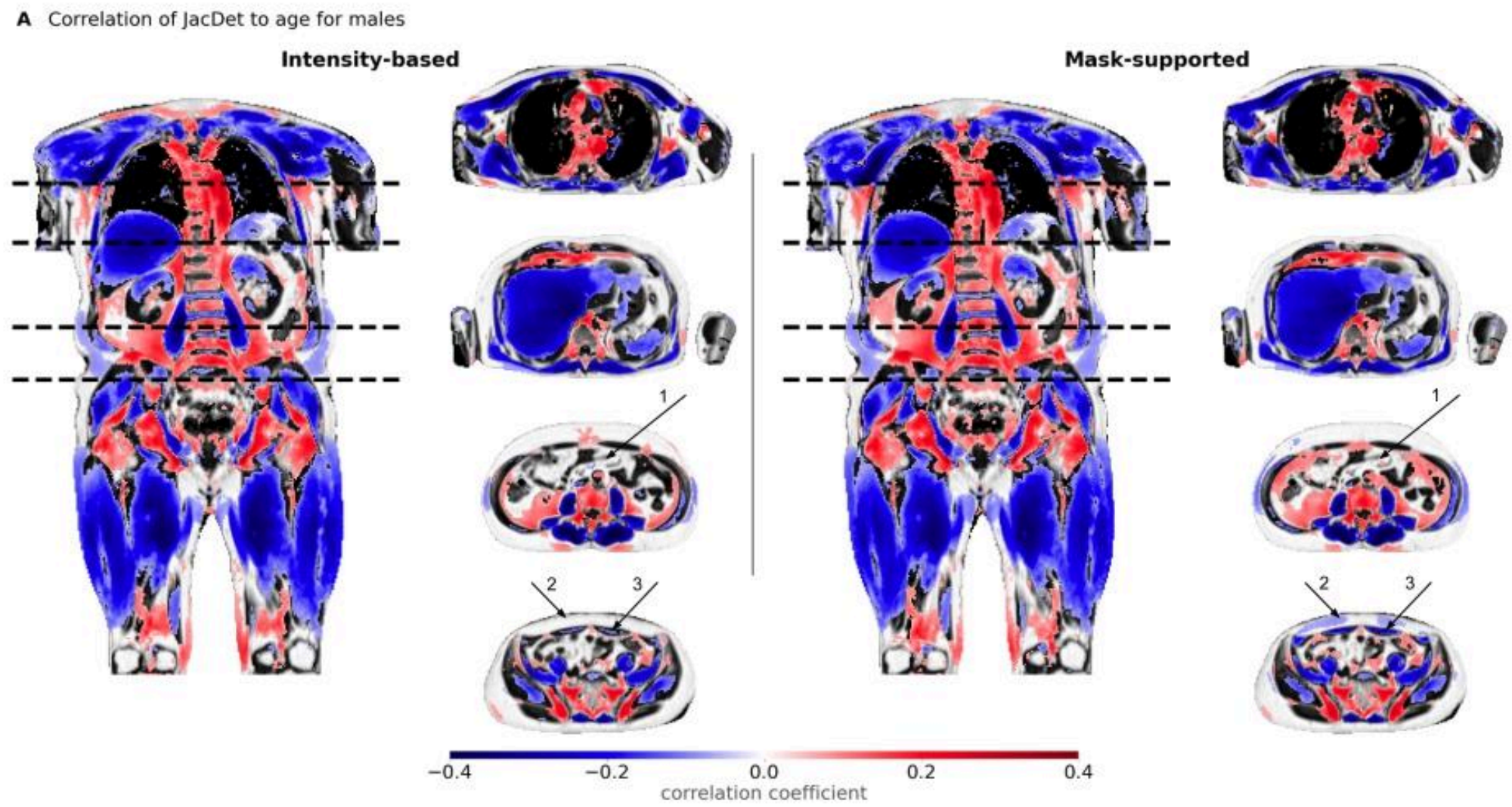

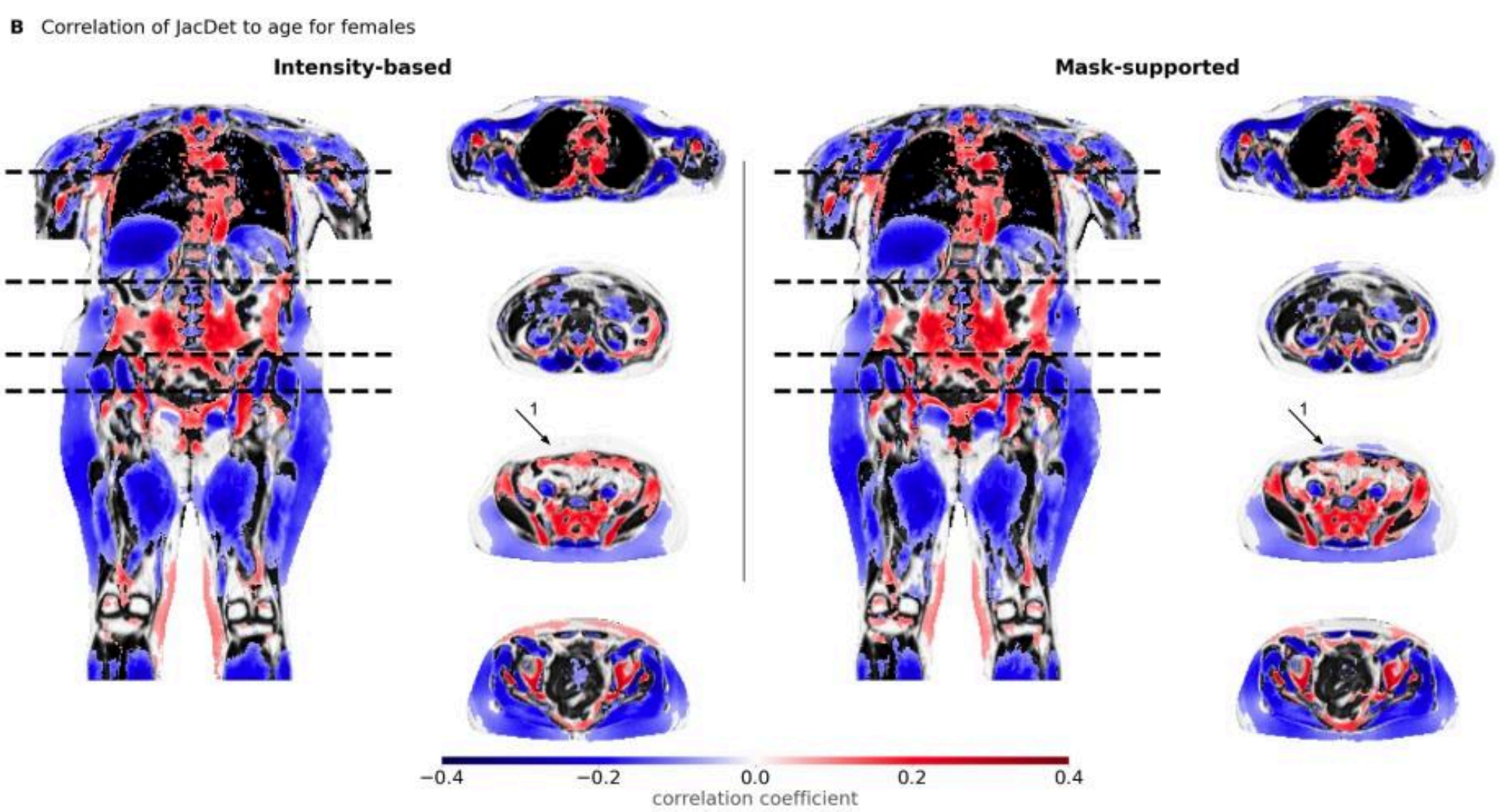

**Fig 10:** Correlation maps showing the voxel-wise Pearson correlation coefficient between age and JD for the intensity-based and mask-supported registrations, for the males (A) and females (B). Only significant correlation coefficients based on FDR-adjusted p-values are displayed. The correlation maps are overlaid on the reference subject FF images. The dotted horizontal lines in the coronal image indicate the position of the axial images. Reproduced by kind permission of UK Biobank ©.

# Discussion and Conclusion

We propose a novel image registration method that uses segmentation masks of SAT and muscle to guide the deformation, with the aim of improving deformable image registration in whole-body water-fat MRI across individuals. To evaluate the proposed mask-supported method, images from 4000 individuals were registered to sex-specific reference images, and the registration performance was compared to that of a similar registration method (intensity-based method), applied to the same dataset but without the support of segmentation masks. To evaluate the mask-supported method against established approaches, its performance was also compared to that achieved with two other previously published registration methods (uniGradICON and MIRTK). The proposed mask-supported registration method outperformed the other methods in terms of tissue label overlap. The improved registration method also led to more pronounced correlations, evaluated both qualitatively and quantitatively, between age and each of FF and JD in abdominal SAT and muscle, demonstrating a potential benefit for large-scale epidemiological imaging studies targeting associations between body composition and disease at the voxel/organ/tissue level.

Along with proposing a mask-supported registration method, we also conducted a full cohort evaluation of registered UK Biobank MR images. 71 organs and tissues were segmented using an AI-based method, VIBESegmentator, and all of these segmentation masks were used for the evaluation, including the SAT and muscle masks used as input for the registration. The proposed mask-supported registration method achieved a mean Dice score of 0.77 and 0.75 across all of the segmented regions for males and females, respectively. In comparison to the intensity-based method, the mask-supported method showed an increase of 6pp in both males and females. When compared to uniGradICON and MIRTK, the increase was 9pp / 8pp and 12pp / 13pp, respectively. Segmentations from two other AI-based methods, MRSegmentator (number of segmented regions = 40) and TotalSegmentator (number of segmented regions = 50), were used for additional evaluation. The mask-supported registration achieved a mean Dice score of 0.736 / 0.676 when evaluated with MRSegmentator and 0.683 / 0.617 when evaluated with TotalSegmentator for males / females, respectively. When comparing the MRSegmentator evaluations of the mask-supported method to the intensity-based method / uniGRADICON / MIRTK, the mean Dice scores were 0.7 pp / 4 pp / 12 pp higher for males and 1 pp / 4 pp / 11 pp higher for females. The evaluation with TotalSegmentator showed improvements of 2 pp / 5 pp / 10 pp for males and 2 pp / 5 pp / 11 pp for females.

Previous studies [13, 16, 17] have shown that including masks in medical image registration tasks improves registration quality in both PET/CT and MR images. Ahmad et. al. and Jönsson et. al. have both used image-wise metrics such as inverse consistency for the evaluation. Jönsson et. al. have also evaluated on a voxel-wise level by applying the registration method to analyze a local volume change map (JD map) for metabolic tumor volumes. However, neither of these studies includes an in-depth tissue or organ-wise analysis to evaluate the benefit of mask-supported registration, such as the Dice score or label error frequency map evaluations in our study.

In this study, the reference subjects were stratified only by sex, showing that the method allows cohort-wise registration of subjects with a wide range of body mass indices (BMIs) and body composition to a median reference subject. Improvements in Dice score and tissue prediction metrics were shown for both the male and female cohorts. The proposed method is successful in registering both sexes using the same methodology and parameters, not requiring further tuning.

Another important consideration is the inclusion of the VIBESegmentator SAT and muscle masks, both in the mask-supported registration and the evaluation. These two masks showed significant improvement in Dice score in comparison to the intensity-based, MIRTK, and uniGradICON models. We believe that it is important to report the Dice scores for all 71 segmented organs and tissues. A majority of the other segmentations also showed significant improvement with the proposed method, proving that the registration was successful in aligning the anatomy also outside of the SAT and muscle. While the regions adjacent to these two tissues, such as bones, might have residual bias due to the inclusion of the two mask channels in the registration, this would not spread to the inner structures, like the liver or kidneys. When presenting the results, SAT and muscle Dice scores were considered separately and were evaluated with the potential bias in mind. Mean Dice scores were presented with and without including these two tissues for the VIBESegmentor-based evaluations. Two other, independent segmentation models were additionally used to segment and evaluate the registrations to present additional, unbiased results.

The main limitation of the method is that it is reliant on the quality of the segmentation masks. The inclusion of the SAT and muscle masks led to improved registration performance, in comparison to the intensity-based method that was not supported by masks. If these masks are not of acceptable accuracy, the registration and evaluation results would likely also be affected. While some regions, such as muscle, fat tissue and bones, were registered successfully, as shown by the high Dice scores across all three models, the same accuracy was not seen in other segments like the pancreas, gallbladder, and small structures such as veins and glands. This should be taken into account for downstream analyses using the registered images.

Another limitation of the study is related to the dataset. Some images did not include the arms, knees, and shoulder regions, due to variations in the field-of-view and subject positioning, and others were of low quality. The 4000 image volumes used in this study were all manually quality controlled, and the images deemed to be of low quality were replaced by images of adequate quality. Under the assumption that the imaging protocol was standardized across time and sites, the image quality of the subset used in this study should be representative of the entire cohort.

An important consideration when analysing the method comparisons is that uniGradICON was not fine-tuned for this specific cohort. In line with its design as an out‑of‑the‑box image registration method, the model was used as provided, using the default configuration and without additional training. For MIRTK, the parameters used by Starck et al. for the same UK Biobank image cohort were applied. No additional training or ablation study was conducted to further optimise the parameters. The reported improvements in registration performance apply to the UK Biobank neck‑to‑knee MR images employed in this study. Further evaluation on additional cohorts, following parameter optimization, is warranted to determine whether superior performance can be achieved with different inputs.

On average, the runtime for a single pair of moving and reference images using the mask-supported method was ~3 minutes, which is the highest among the four methods that were compared in this study. This could be reduced by limiting the number of iterations per resolution or stopping at the lower resolution step, although this was found to reduce the quality in preliminary tests.

A critical aspect of the image registration is the selection of reference subjects. The different anatomies of males and females make it challenging to register images of both sexes to a common reference image. For the proposed method, one reference subject was chosen for each sex, and all the analyses were conducted separately for males and females. However, a method that can successfully register all subjects to a single reference, regardless of sex, would allow for a common analysis of the entire cohort, like voxel-wise studies of sex differences in associations. Selection of the reference subjects can also lead to potential bias. A small-scale sensitivity analysis using reference subjects within different BMI categories indicated that the proposed method was relatively robust to the choice of reference subject. A possible way to avoid reference subject-induced bias would be to create a synthetic space to register all subject images to, eliminating the need for selecting sex-stratified references.

Potential future improvements for the proposed method are increasing the number of masks used to guide the registration and decreasing the runtime for individual image pairs. The runtime could be reduced by stopping at an earlier resolution pyramid level or using a lower number of iterations per level, thus sacrificing some performance. Additional segmentation masks of smaller specific structures, instead of general and larger tissue masks, would allow for more detailed guidance of the registration. This could also decrease the runtime by guiding the lower levels of the pyramid better and reducing the number of iterations necessary in higher levels. Different combinations of segmentation masks can be tested to evaluate the change in performance. It is also possible to develop more efficient ways to include segmentation masks in deep learning methods in order to combine the proposed method with the previously published and existing deep learning methods such as uniGradICON and MIRTK.

Cohort-wise registrations can be used for voxel-wise studies on disease and risk factors, integrating non-imaging data from UK Biobank, multi-omics, including age, and longitudinal studies for subjects who have completed their follow-up scans.

In conclusion, we present and evaluate a whole-body image registration method that uses two tissue masks to guide the registration. This mask-supported approach was evaluated against the same method without including the segmentation masks, as well as two previously published methods, uniGradICON and MIRTK. The proposed method showed higher performance in terms of voxel-wise statistics like Dice scores and label error frequency maps, as well as a lower frequency of JD folding when compared to the intensity-based method. The correlation maps of age and each of FF and JD were also observed to be sharper. This result showed that the addition of segmentation masks improved registration performance for neck-to-knee Dixon MR images. Further analysis is required to determine if this effect can be replicated using other registration methods, cohorts, or modalities.

## Data Availability

The UK Biobank dataset can be accessed through the UK Biobank Access Management Portal (https://ams.ukbiobank.ac.uk/ams/) after registration and approval. The used image data has the field ID 20201.

## Code Availability

The registration method deform can be accessed through GitHub (https://github.com/simeks/deform). All used parameters were specified in the methods section of this paper.

## Acknowledgements

The computations and data handling used resources provided by the Swedish National Infrastructure for Computing (SNIC) at Uppsala University, partially funded by the Swedish Research Council through grant agreement no. 2018-05973.

This research has been conducted using the UK Biobank Resource under Application Number 14237 and the Swedish National Infrastructure for Computing (SNIC, sens2019016).

## Funding Information

This research was funded by grants from the Swedish Heart-Lung Foundation (20240402) and the Swedish Research Council (2016-01040, 2019-04756, 2023-03607), EXODIAB, and Uppsala Diabetes Center (UDC).

## Author Contributions

J.K, J.Ö. and Y.U. designed the study. Y.U. conducted all experiments, analyses, and wrote the first draft of the manuscript. J.K., E.L., H.A., and J.Ö. supervised the study. All authors contributed to discussing the results and reviewing the manuscript.

## Competing Interests

J.K and H.A are co-founders and part-time employees of Antaros Medical AB.

# Supplementary Information

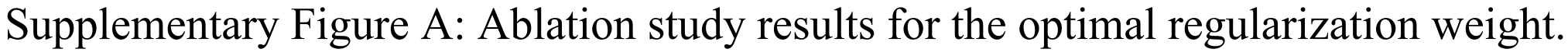

Supplementary Figure A: Ablation study results for the optimal regularization weight.

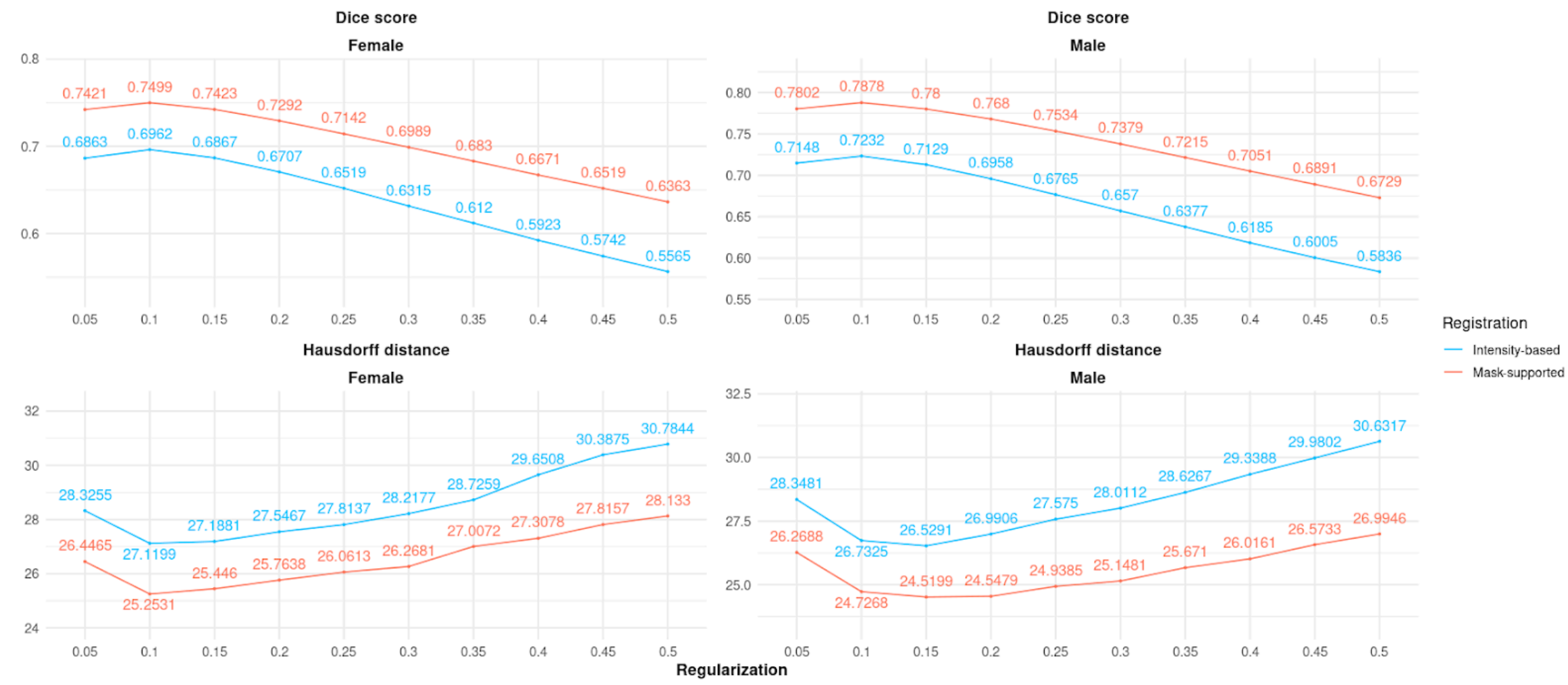

Mean dice scores and Hausdorff distances across the 71 masks and all subjects for different regularization weights for females and males. For the mask-supported registrations, the weight of the mask channel was set to 0.6 for these experiments.

Supplementary Figure B: Ablation study results to obtain the optimal weight for the SAT and muscle weights.

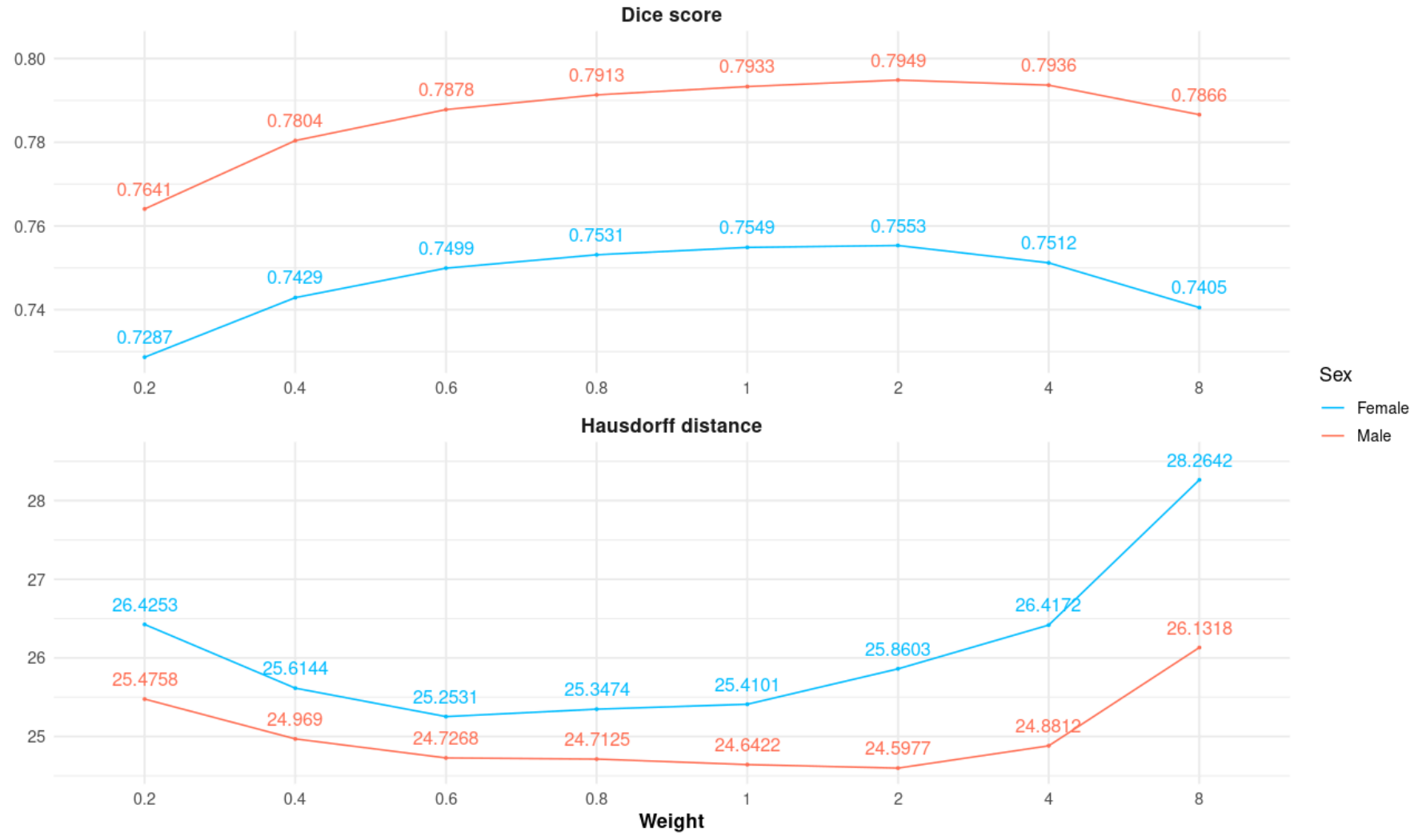

Mean dice scores and Hausdorff distances across the 71 masks and all subjects for different binary mask input channel weights for the mask-supported registration. The regularization weight for these experiments was set to 0.1.

Supplementary Figure C: Mean FF images for males and females registered with different mask channel weights

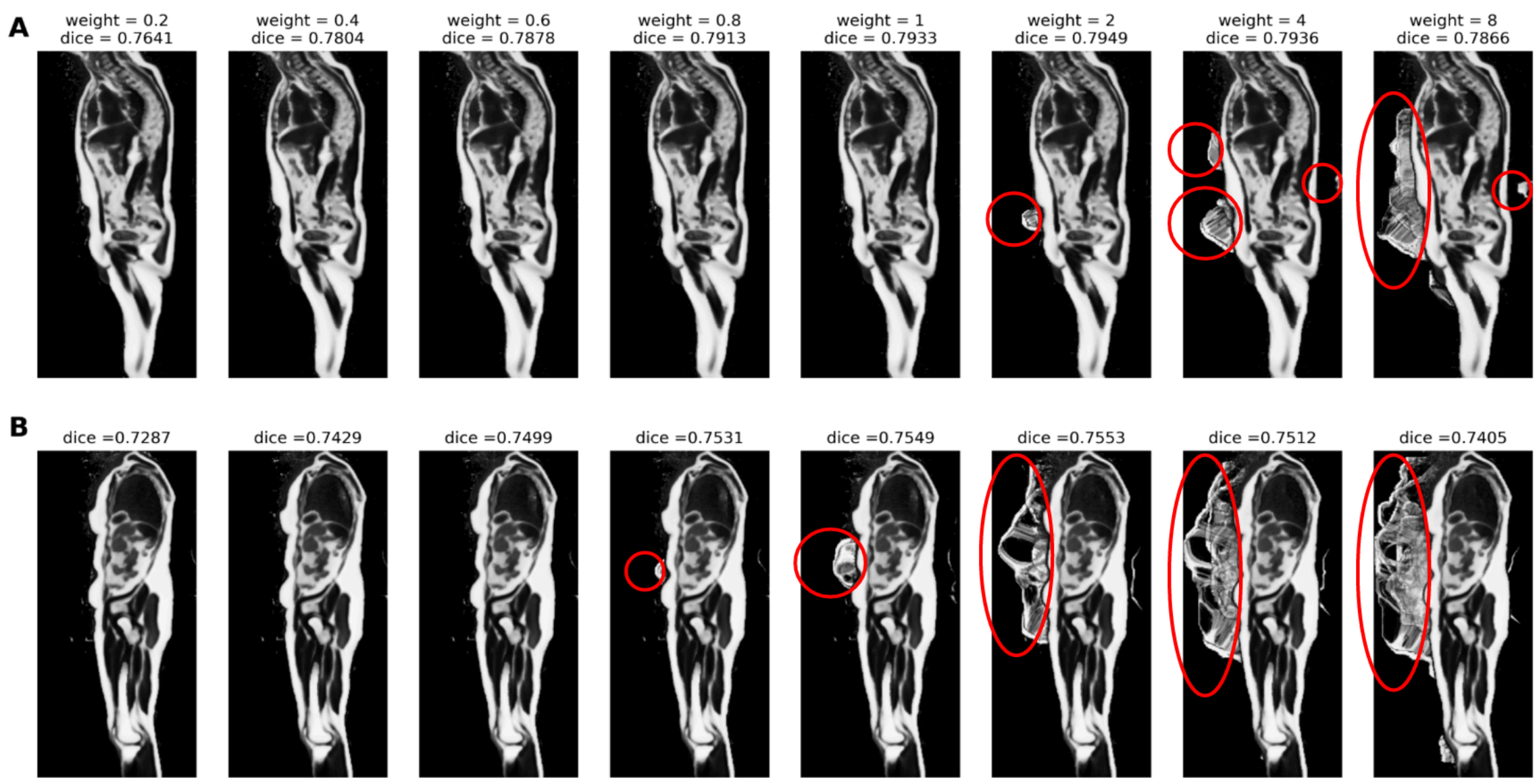

Mean FF images for the 200 males and females of the ablation sub-cohort that were registered using different mask channel weights. All registrations used 0.1 regularization. The observed artifacts have been circled. The mean images were calculated by using the number of subjects that have non-zero values for the voxel. Reproduced by kind permission of UK Biobank ©.

Supplementary Figure D: Mean number of voxels outside of the body mask in the deformed FF images.

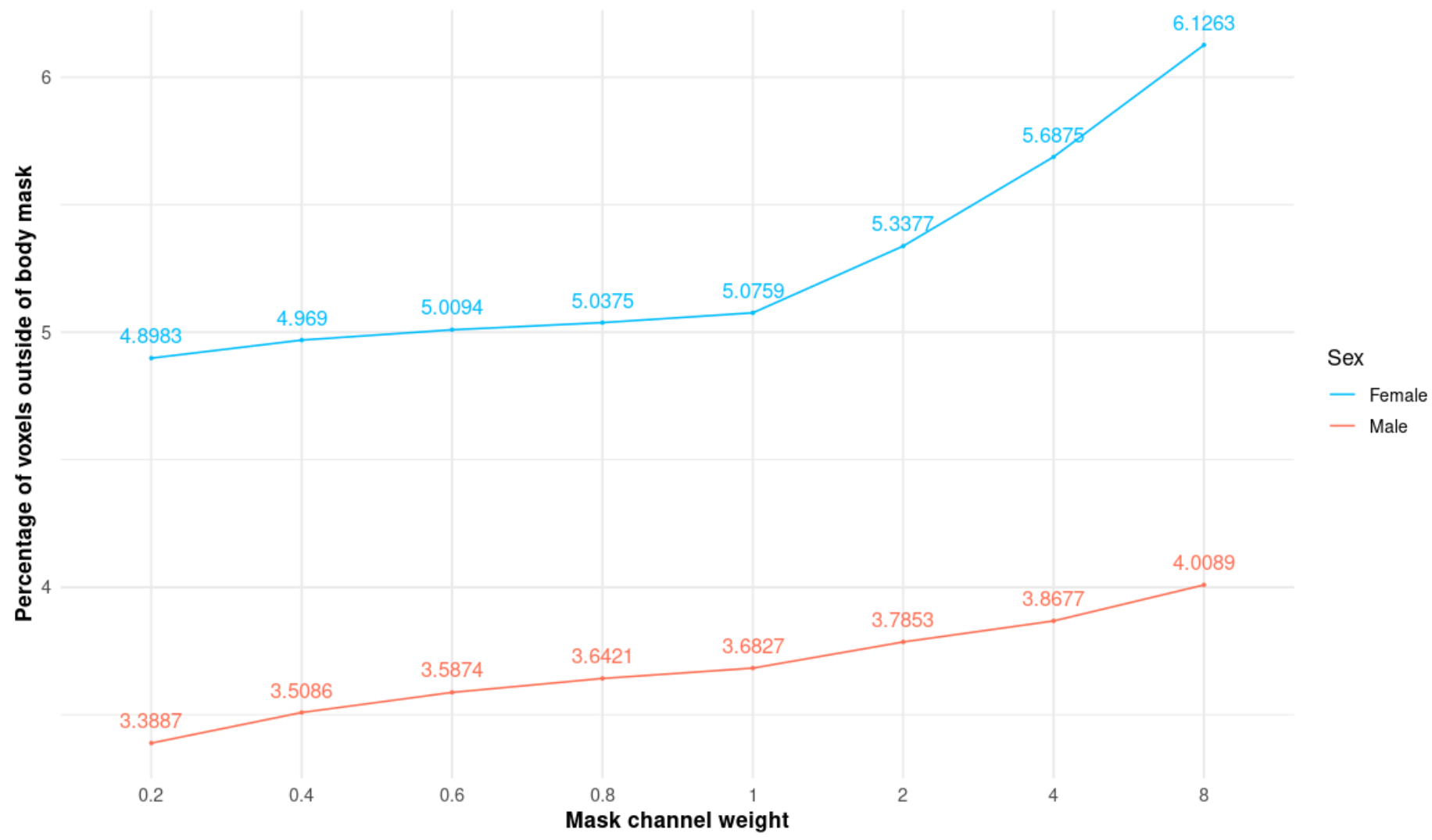

Mean number of voxels outside of the reference subject's bodymask in mask-supported registration. Results are presented for the 200 subjects in the ablation subset, separately for males and females.

Supplementary Figure E: Per-voxel mean Jacobian Determinant fold rate maps for males and females with the intensity-based and mask-supported registrations.

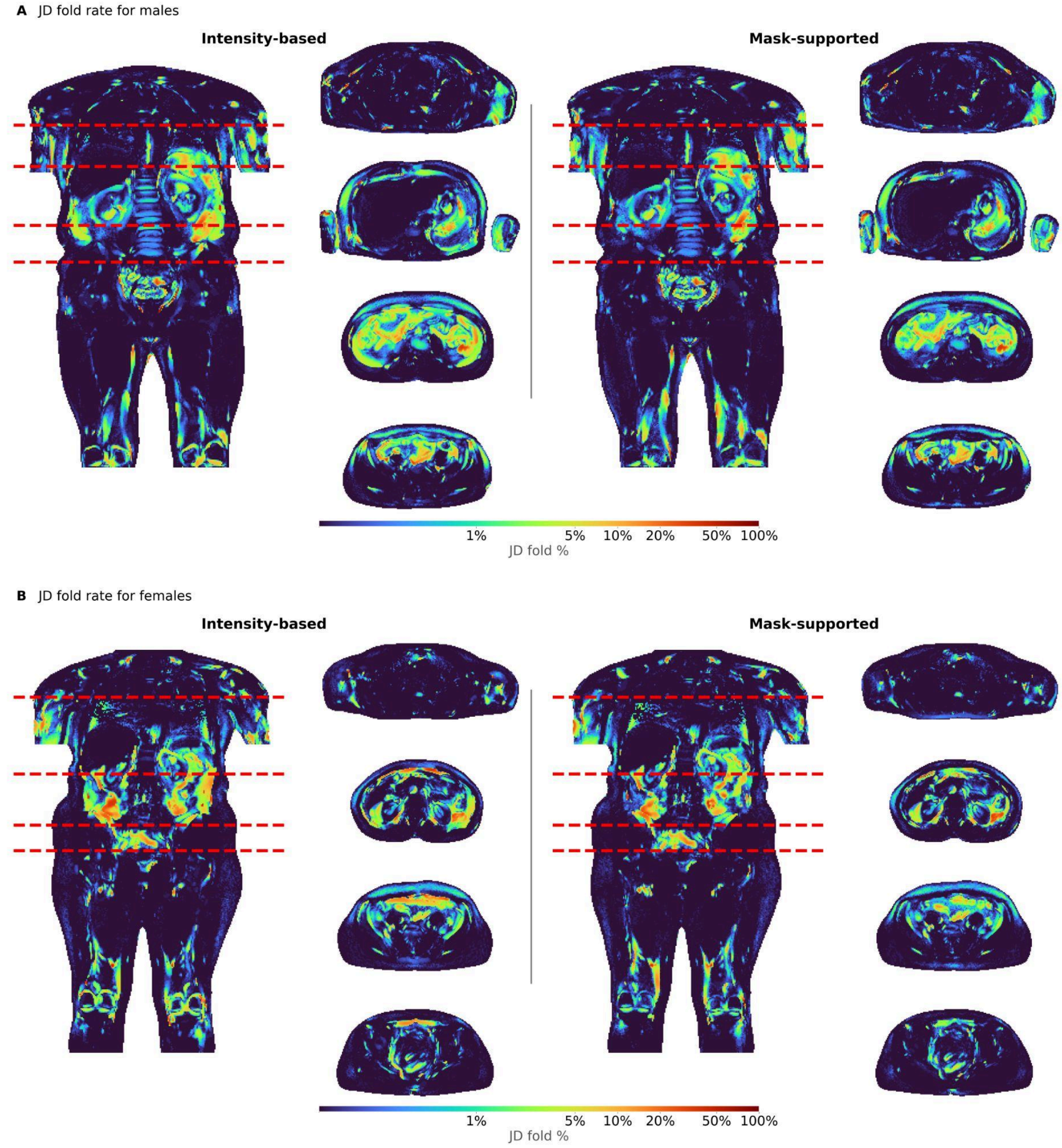

JD fold frequencies for the 2000 subjects per-sex and registration method. The colormap was scaled logarithmically for ease of visualisation. Reproduced by kind permission of UK Biobank ©.

Supplementary Table A: Parameters used for the mask-supported registration.

| Parameter | Value |
|---|---|
| pyramid levels | 6 |
| pyramid stop level | 0 |
| block size | [12, 12, 12] |
| block energy epsilon | 1.00E-07 |
| max iteration count | 100 |
| step size | 0.5 |
| regularization scale | 1 |
| regularization exponent | 2 |
| regularization weight | 0.1 |
| image resampler | gaussian |
| cost function | ssd |
| update rule | additive |
| image normalization | TRUE |
| FF / WF image weight | 1 |
| binary mask weight | 0.6 |

Supplementary Table B: Demographic information of the 4000 subjects in this analysis and the reference subjects for each sex.

| | **male** | | **female** | |
|---|---|---|---|---|
| | **cohort mean** | **reference subject** | **cohort mean** | **reference subject** |
| **age** | 64.7898 | 64.00 | 63.6863 | 63.59 |
| **body mass index (kg/m2)** | 27.1984 | 27.90 | 26.0847 | 22.70 |
| **waist circumference (cm)** | 94.7993 | 98 | 81.7322 | 73 |
| **hip circumference (cm)** | 102.8402 | 104 | 101.7272 | 98 |
| **body surface area (m2)** | 1.9978 | 1.97 | 1.7329 | 1.63 |

Supplementary Table C: Mean dice scores of the 71 masks segmented with VIBESegmentator for the female and male cohorts for the intensity-based and mask-supported registrations. The statistical tests were done using pairwise Wilcoxon signed-rank tests with adjustment for multiple testing using Bonferroni correction. (na: not applicable, ns: not significant, *: p-value < 0.05, **: p-value < 0.01, ***: p-value < 0.001, ****: p-value < 0.0001)

| | **male** | | | | **female** | | | |
|---|---|---|---|---|---|---|---|---|
| **mask name** | **intensity-based** | **mask-supported** | **p.adj** | **significance** | **intensity-based** | **mask-supported** | **p.adj** | **significance** |
| **adrenal gland left** | 0.5632 | 0.5631 | 1.000 | ns | **0.3459** | 0.3380 | 0.000 | **** |
| **adrenal gland right** | 0.5567 | 0.5573 | 1.000 | ns | **0.3372** | 0.3292 | 0.000 | **** |
| **aorta** | 0.8467 | 0.8466 | 1.000 | ns | 0.7603 | **0.7649** | 0.000 | **** |
| **atrial appendage left** | **0.6055** | 0.5988 | 0.000 | **** | 0.3926 | 0.3924 | 1.000 | ns |
| **autochthon left** | 0.9179 | **0.9514** | 0.000 | **** | 0.9056 | **0.9397** | 0.000 | **** |
| **autochthon right** | 0.9097 | **0.9526** | 0.000 | **** | 0.9074 | **0.9468** | 0.000 | **** |
| **bone other** | 0.4908 | **0.6070** | 0.000 | **** | 0.6903 | **0.8310** | 0.000 | **** |
| **brachiocephalic trunk** | 0.6492 | **0.6817** | 0.000 | **** | 0.6156 | **0.6278** | 0.000 | **** |
| **brachiocephalic vein left** | 0.6300 | **0.6774** | 0.000 | **** | 0.6352 | **0.6653** | 0.000 | **** |
| **brachiocephalic vein right** | 0.6289 | **0.7202** | 0.000 | **** | 0.7224 | **0.7340** | 0.000 | **** |
| **clavicula left** | 0.6171 | **0.8372** | 0.000 | **** | 0.5898 | **0.8031** | 0.000 | **** |
| **clavicula right** | 0.4781 | **0.8131** | 0.000 | **** | 0.6494 | **0.8541** | 0.000 | **** |
| **common carotid artery left** | 0.5517 | **0.7214** | 0.000 | **** | 0.4978 | **0.6460** | 0.000 | **** |
| **common carotid artery right** | 0.4363 | **0.7618** | 0.000 | **** | 0.5734 | **0.7743** | 0.000 | **** |
| **costal cartilages** | 0.5480 | **0.7067** | 0.000 | **** | 0.6387 | **0.7535** | 0.000 | **** |
| **duodenum** | 0.6427 | **0.6610** | 0.000 | **** | 0.4560 | **0.4641** | 0.000 | **** |
| **esophagus** | 0.7012 | **0.7396** | 0.000 | **** | 0.6724 | **0.6969** | 0.000 | **** |
| **femur left** | 0.8903 | **0.9281** | 0.000 | **** | 0.8932 | **0.9656** | 0.000 | **** |
| **femur right** | 0.8879 | **0.9042** | 0.000 | **** | 0.8885 | **0.9602** | 0.000 | **** |
| **gallbladder** | 0.3550 | **0.3624** | 0.000 | **** | 0.2107 | **0.2199** | 0.000 | **** |
| **gluteus maximus left** | 0.9297 | **0.9651** | 0.000 | **** | 0.9325 | **0.9704** | 0.000 | **** |
| **gluteus maximus right** | 0.9438 | **0.9662** | 0.000 | **** | 0.9332 | **0.9717** | 0.000 | **** |
| **gluteus medius left** | 0.8926 | **0.9197** | 0.000 | **** | 0.8827 | **0.9155** | 0.000 | **** |
| **gluteus medius right** | 0.8999 | **0.9184** | 0.000 | **** | 0.8786 | **0.9124** | 0.000 | **** |
| **gluteus minimus left** | 0.8488 | **0.8678** | 0.000 | **** | 0.8213 | **0.8512** | 0.000 | **** |
| **gluteus minimus right** | 0.8451 | **0.8589** | 0.000 | **** | 0.8235 | **0.8569** | 0.000 | **** |
| **heart** | **0.8876** | 0.8876 | 0.000 | **** | **0.8961** | 0.8947 | 0.000 | **** |
| **hip left** | 0.8625 | **0.8991** | 0.000 | **** | 0.8562 | **0.9053** | 0.000 | **** |
| **hip right** | 0.8818 | **0.9015** | 0.000 | **** | 0.8423 | **0.8975** | 0.000 | **** |
| **humerus left** | 0.8061 | **0.9283** | 0.000 | **** | 0.7408 | **0.9209** | 0.000 | **** |
| **humerus right** | 0.5791 | **0.8026** | 0.000 | **** | 0.7917 | **0.9265** | 0.000 | **** |
| **iliac artery left** | 0.5562 | **0.5825** | 0.000 | **** | 0.5802 | **0.5986** | 0.000 | **** |
| **iliac artery right** | 0.5087 | **0.5375** | 0.000 | **** | 0.5517 | **0.5890** | 0.000 | **** |
| **iliac vena left** | 0.6843 | **0.6999** | 0.000 | **** | 0.6840 | **0.7021** | 0.000 | **** |
| **iliac vena right** | 0.6117 | **0.6239** | 0.000 | **** | 0.6444 | **0.6669** | 0.000 | **** |
| **iliopsoas left** | 0.8848 | **0.8999** | 0.000 | **** | 0.8444 | **0.8637** | 0.000 | **** |
| **iliopsoas right** | 0.8808 | **0.8916** | 0.000 | **** | 0.8163 | **0.8403** | 0.000 | **** |
| **inferior vena cava** | **0.7368** | 0.7339 | 0.000 | **** | 0.5960 | **0.5991** | 0.000 | *** |
| **inner fat** | 0.7302 | **0.7684** | 0.000 | **** | 0.6303 | **0.6910** | 0.000 | **** |
| **intestine** | 0.7371 | **0.7818** | 0.000 | **** | 0.7489 | **0.7963** | 0.000 | **** |
| **IVD** | 0.6588 | **0.6615** | 0.000 | **** | 0.5405 | **0.5463** | 0.000 | **** |
| **kidney left** | 0.8531 | **0.8636** | 0.000 | **** | **0.7547** | 0.7452 | 0.000 | **** |
| **kidney right** | 0.8708 | **0.8753** | 0.000 | **** | 0.6965 | 0.6969 | 1.000 | ns |
| **liver** | 0.9103 | **0.9143** | 0.000 | **** | 0.8677 | **0.8722** | 0.000 | **** |
| **lung lower lobe left** | 0.8317 | **0.8343** | 0.000 | **** | 0.8960 | **0.9018** | 0.000 | **** |
| **lung lower lobe right** | 0.8466 | **0.8468** | 0.000 | **** | 0.8796 | **0.8858** | 0.000 | **** |
| **lung middle lobe right** | 0.8250 | **0.8313** | 0.000 | **** | 0.8284 | **0.8369** | 0.000 | **** |
| **lung upper lobe left** | 0.8963 | **0.9092** | 0.000 | **** | 0.9038 | **0.9171** | 0.000 | **** |
| **lung upper lobe right** | 0.8813 | **0.8924** | 0.000 | **** | 0.9022 | **0.9121** | 0.000 | **** |
| **muscle** | 0.8867 | **0.9647** | 0.000 | **** | 0.8769 | **0.9658** | 0.000 | **** |
| **pancreas** | 0.6206 | **0.6298** | 0.000 | **** | 0.3445 | **0.3483** | 0.000 | *** |
| **portal vein and splenic vein** | 0.5514 | **0.5564** | 0.000 | **** | **0.2840** | 0.2775 | 0.000 | **** |
| **prostate** | 0.6197 | **0.6759** | 0.000 | **** | NA | NA | NA | NA |
| **pulmonary vein** | 0.6227 | 0.6217 | 1.000 | ns | **0.6088** | 0.6068 | 0.014 | * |
| **sacrum** | 0.8850 | **0.8963** | 0.000 | **** | 0.8374 | **0.8592** | 0.000 | **** |
| **scapula left** | 0.7844 | **0.9411** | 0.000 | **** | 0.6673 | **0.8947** | 0.000 | **** |
| **scapula right** | 0.6238 | **0.8781** | 0.000 | **** | 0.7050 | **0.9077** | 0.000 | **** |
| **spinal channel** | 0.7337 | **0.7500** | 0.000 | **** | 0.7138 | **0.7259** | 0.000 | **** |
| **spinal cord** | 0.6449 | **0.6762** | 0.000 | **** | 0.6507 | **0.6697** | 0.000 | **** |
| **spleen** | 0.7001 | **0.7064** | 0.000 | *** | 0.7005 | 0.6985 | 0.140 | ns |
| **sternum** | 0.6111 | **0.7286** | 0.000 | **** | 0.7636 | **0.8474** | 0.000 | **** |
| **stomach** | **0.6393** | 0.6361 | 0.003 | ** | 0.4958 | **0.5070** | 0.000 | **** |
| **subclavian artery left** | 0.5770 | **0.7504** | 0.000 | **** | 0.5259 | **0.6708** | 0.000 | **** |
| **subclavian artery right** | 0.5567 | **0.7990** | 0.000 | **** | 0.6519 | **0.7993** | 0.000 | **** |
| **subcutaneous fat** | 0.8795 | **0.9628** | 0.000 | **** | 0.9169 | **0.9756** | 0.000 | **** |
| **superior vena cava** | 0.7209 | **0.7281** | 0.000 | **** | 0.7825 | 0.7826 | 0.280 | ns |
| **thyroid gland** | 0.4876 | **0.8294** | 0.000 | **** | 0.5558 | **0.8086** | 0.000 | **** |
| **trachea** | 0.7283 | **0.7985** | 0.000 | **** | 0.7058 | **0.7455** | 0.000 | **** |
| **urinary bladder** | 0.6281 | **0.6433** | 0.000 | **** | 0.6975 | **0.7086** | 0.000 | **** |
| **vertebra body** | 0.7925 | **0.8001** | 0.000 | **** | 0.7165 | **0.7252** | 0.000 | **** |
| **vertebra posterior elements** | 0.6254 | **0.6495** | 0.000 | **** | 0.6145 | **0.6329** | 0.000 | **** |

Supplementary Table D: Mean dice scores of the 40 masks segmented with MRSegmentator for the female and male cohorts for the mask-supported and intensity-based registrations. The statistical tests were done using pairwise Wilcoxon signed-rank tests with adjustment for multiple testing using Bonferroni correction. (na: not applicable, ns: not significant, *: p-value < 0.05, **: p-value < 0.01, ***: p-value < 0.001, ****: p-value < 0.0001)

| | **male** | | | | **female** | | | |
|---|---|---|---|---|---|---|---|---|
| **mask name** | **intensity based** | **mask supported** | **p.adj** | **significance** | **intensity based** | **mask supported** | **p.adj** | **significance** |
| **aorta** | 0.8476 | **0.8476** | 0.3600 | ns | 0.7662 | **0.7701** | 0.0000 | **** |
| **colon** | 0.4827 | **0.5122** | 0.0000 | **** | 0.4050 | **0.4312** | 0.0000 | **** |
| **duodenum** | 0.5458 | **0.5616** | 0.0000 | **** | 0.4248 | **0.4316** | 0.0000 | **** |
| **esophagus** | 0.6512 | **0.6550** | 0.0000 | **** | 0.5673 | **0.5689** | 0.0031 | ** |
| **gallbladder** | 0.2457 | **0.2485** | 0.0024 | ** | 0.1119 | **0.1198** | 0.0000 | **** |
| **heart** | **0.8883** | 0.8880 | 0.3600 | ns | **0.8850** | 0.8835 | 0.0000 | **** |
| **inferior_vena_cava** | **0.6774** | 0.6760 | 0.0064 | ** | 0.4957 | **0.4965** | 1.0000 | ns |
| **left_adrenal_gland** | 0.5163 | **0.5170** | 1.0000 | ns | NA | NA | NA | NA |
| **left_autochthonous_muscle** | 0.9073 | **0.9102** | 0.0000 | **** | 0.8973 | **0.9017** | 0.0000 | **** |
| **left_femur** | 0.8686 | **0.8838** | 0.0000 | **** | 0.8844 | **0.9151** | 0.0000 | **** |
| **left_gluteus_maximus** | 0.9182 | **0.9234** | 0.0000 | **** | 0.9182 | **0.9247** | 0.0000 | **** |
| **left_gluteus_medius** | 0.8793 | **0.8878** | 0.0000 | **** | 0.8798 | **0.8910** | 0.0000 | **** |
| **left_gluteus_minimus** | 0.7949 | **0.8054** | 0.0000 | **** | 0.7623 | **0.7720** | 0.0000 | **** |
| **left_hip** | 0.8459 | **0.8676** | 0.0000 | **** | 0.8405 | **0.8706** | 0.0000 | **** |
| **left_iliac_artery** | 0.4833 | **0.4966** | 0.0000 | **** | 0.4802 | **0.4952** | 0.0000 | **** |
| **left_iliac_vena** | 0.6463 | **0.6530** | 0.0000 | **** | 0.6705 | **0.6830** | 0.0000 | **** |
| **left_iliopsoas_muscle** | 0.8837 | **0.8873** | 0.0000 | **** | 0.8444 | **0.8490** | 0.0000 | **** |
| **left_kidney** | 0.8567 | **0.8672** | 0.0000 | **** | **0.7560** | 0.7473 | 0.0000 | **** |
| **left_lung** | 0.9076 | **0.9097** | 0.0000 | **** | 0.9358 | **0.9390** | 0.0000 | **** |
| **liver** | 0.9069 | **0.9091** | 0.0000 | **** | 0.8661 | **0.8673** | 0.0000 | **** |
| **pancreas** | 0.5188 | **0.5279** | 0.0000 | **** | 0.2971 | **0.3001** | 0.0301 | * |
| **portal_vein_and_splenic_vein** | 0.3672 | **0.3713** | 0.0000 | **** | **0.1505** | 0.1479 | 0.0115 | * |
| **right_adrenal_gland** | 0.4968 | **0.4989** | 0.0035 | ** | **0.2215** | 0.2147 | 0.0000 | **** |
| **right_autochthonous_muscle** | 0.8925 | **0.8982** | 0.0000 | **** | 0.8817 | **0.8869** | 0.0000 | **** |
| **right_femur** | 0.8662 | **0.8798** | 0.0000 | **** | 0.8821 | **0.9110** | 0.0000 | **** |
| **right_gluteus_maximus** | 0.9331 | **0.9360** | 0.0000 | **** | 0.9177 | **0.9260** | 0.0000 | **** |
| **right_gluteus_medius** | 0.9006 | **0.9046** | 0.0000 | **** | 0.8767 | **0.8895** | 0.0000 | **** |
| **right_gluteus_minimus** | 0.8156 | **0.8237** | 0.0000 | **** | 0.7751 | **0.7881** | 0.0000 | **** |
| **right_hip** | 0.8632 | **0.8763** | 0.0000 | **** | 0.8257 | **0.8613** | 0.0000 | **** |
| **right_iliac_artery** | 0.3998 | **0.4013** | 0.0000 | **** | 0.4658 | **0.4856** | 0.0000 | **** |
| **right_iliac_vena** | 0.5399 | **0.5453** | 0.0000 | **** | 0.5766 | **0.5931** | 0.0000 | **** |
| **right_iliopsoas_muscle** | 0.8784 | **0.8819** | 0.0000 | **** | 0.8280 | **0.8372** | 0.0000 | **** |
| **right_kidney** | 0.8776 | **0.8824** | 0.0000 | **** | 0.6869 | **0.6877** | 1.0000 | ns |

| **right_lung** | 0.9208 | **0.9223** | 0.0000 | **** | 0.9380 | **0.9406** | 0.0000 | **** |
|---|---|---|---|---|---|---|---|---|
| **sacrum** | 0.8356 | **0.8482** | 0.0000 | **** | 0.7874 | **0.8101** | 0.0000 | **** |
| **small_bowel** | 0.5310 | **0.5639** | 0.0000 | **** | 0.4897 | **0.5325** | 0.0000 | **** |
| **spine** | 0.7827 | **0.7831** | 1.0000 | ns | 0.6902 | **0.6946** | 0.0000 | **** |
| **spleen** | 0.7116 | **0.7186** | 0.0007 | *** | **0.7062** | 0.7002 | 0.0000 | **** |
| **stomach** | **0.6022** | 0.5982 | 0.0000 | **** | 0.4507 | **0.4616** | 0.0000 | **** |
| **urinary_bladder** | 0.6120 | **0.6291** | 0.0000 | **** | 0.6550 | **0.6687** | 0.0000 | **** |

Supplementary Table E: Mean dice scores of the 50 masks segmented with TotalSegmentator for the female and male cohorts for the mask-supported and intensity-based registrations. The statistical tests were done using pairwise Wilcoxon signed-rank tests with adjustment for multiple testing using Bonferroni correction. (na: not applicable, ns: not significant, *: p-value < 0.05, **: p-value < 0.01, ***: p-value < 0.001, ****: p-value < 0.0001)

| | **male** | | | | **female** | | | |
|---|---|---|---|---|---|---|---|---|
| **mask name** | **intensity based** | **mask supported** | **p.adj** | **significance** | **intensity based** | **mask supported** | **p.adj** | **significance** |
| **adrenal_gland_left** | 0.4327 | **0.4339** | 1.0000 | ns | **0.1495** | 0.1384 | 0.0000 | **** |
| **adrenal_gland_right** | 0.4497 | **0.4516** | 0.2450 | ns | **0.2114** | 0.2034 | 0.0000 | **** |
| **aorta** | **0.8026** | 0.8005 | 0.0000 | **** | 0.7021 | **0.7062** | 0.0000 | **** |
| **autochthon_left** | 0.9039 | **0.9068** | 0.0000 | **** | 0.8885 | **0.8929** | 0.0000 | **** |
| **autochthon_right** | 0.8925 | **0.8994** | 0.0000 | **** | 0.8692 | **0.8762** | 0.0000 | **** |
| **clavicula_left** | 0.4268 | **0.5717** | 0.0000 | **** | 0.3863 | **0.5150** | 0.0000 | **** |
| **clavicula_right** | 0.2816 | **0.5050** | 0.0000 | **** | 0.4156 | **0.5156** | 0.0000 | **** |
| **colon** | 0.4465 | **0.4759** | 0.0000 | **** | 0.3477 | **0.3747** | 0.0000 | **** |
| **duodenum** | 0.4581 | **0.4763** | 0.0000 | **** | 0.2544 | **0.2574** | 0.0000 | **** |
| **esophagus** | 0.6122 | **0.6156** | 0.0000 | **** | 0.5426 | **0.5442** | 0.0480 | * |
| **femur_left** | 0.8563 | **0.8722** | 0.0000 | **** | 0.8665 | **0.8972** | 0.0000 | **** |
| **femur_right** | 0.8619 | **0.8758** | 0.0000 | **** | 0.8785 | **0.9038** | 0.0000 | **** |
| **gallbladder** | 0.2245 | **0.2269** | 0.0165 | * | 0.1133 | **0.1199** | 0.0000 | **** |
| **gluteus_maximus_left** | 0.9100 | **0.9145** | 0.0000 | **** | 0.9125 | **0.9189** | 0.0000 | **** |
| **gluteus_maximus_right** | 0.9149 | **0.9178** | 0.0000 | **** | 0.8996 | **0.9069** | 0.0000 | **** |
| **gluteus_medius_left** | 0.8194 | **0.8269** | 0.0000 | **** | 0.7500 | **0.7589** | 0.0000 | **** |
| **gluteus_medius_right** | 0.8406 | **0.8455** | 0.0000 | **** | 0.7338 | **0.7458** | 0.0000 | **** |
| **gluteus_minimus_left** | 0.7604 | **0.7703** | 0.0000 | **** | 0.7090 | **0.7179** | 0.0000 | **** |
| **gluteus_minimus_right** | 0.7546 | **0.7644** | 0.0000 | **** | 0.7065 | **0.7219** | 0.0000 | **** |
| **heart** | **0.8683** | 0.8678 | 1.0000 | ns | **0.8643** | 0.8632 | 0.0000 | **** |
| **hip_left** | 0.8075 | **0.8251** | 0.0000 | **** | 0.7815 | **0.8027** | 0.0000 | **** |
| **hip_right** | 0.8201 | **0.8312** | 0.0000 | **** | 0.7665 | **0.7971** | 0.0000 | **** |
| **humerus_left** | 0.7820 | **0.8263** | 0.0000 | **** | 0.7075 | **0.8010** | 0.0000 | **** |
| **humerus_right** | 0.5584 | **0.6805** | 0.0000 | **** | 0.7675 | **0.8181** | 0.0000 | **** |
| **iliac_artery_left** | 0.3945 | **0.4032** | 0.0000 | **** | 0.3190 | **0.3290** | 0.0000 | **** |
| **iliac_artery_right** | **0.3400** | 0.3359 | 0.0058 | ** | 0.3420 | **0.3546** | 0.0000 | **** |
| **iliac_vena_left** | 0.5512 | **0.5562** | 0.0000 | **** | 0.5486 | **0.5605** | 0.0000 | **** |
| **iliac_vena_right** | 0.4998 | **0.5024** | 1.0000 | ns | 0.4866 | **0.4995** | 0.0000 | **** |
| **iliopsoas_left** | 0.8518 | **0.8563** | 0.0000 | **** | 0.8147 | **0.8243** | 0.0000 | **** |
| **iliopsoas_right** | 0.8596 | **0.8666** | 0.0000 | **** | 0.8005 | **0.8114** | 0.0000 | **** |
| **inferior_vena_cava** | **0.5731** | 0.5723 | 1.0000 | ns | 0.2340 | **0.2391** | 0.0000 | **** |
| **intervertebral_discs** | 0.6163 | **0.6174** | 1.0000 | ns | 0.5007 | **0.5059** | 0.0000 | **** |
| **kidney_left** | 0.8121 | **0.8224** | 0.0000 | **** | **0.6815** | 0.6746 | 0.0000 | **** |
| **kidney_right** | 0.8250 | **0.8291** | 0.0000 | **** | **0.6057** | 0.6057 | 0.0270 | * |
| **liver** | 0.8903 | **0.8918** | 0.0000 | **** | **0.8500** | 0.8498 | 0.1440 | ns |
| **lung_left** | 0.9060 | **0.9086** | 0.0000 | **** | 0.9327 | **0.9362** | 0.0000 | **** |
| **lung_right** | 0.9212 | **0.9229** | 0.0000 | **** | 0.9415 | **0.9445** | 0.0000 | **** |
| **pancreas** | 0.3579 | **0.3653** | 0.0000 | **** | **0.0608** | 0.0588 | 0.0960 | ns |
| **portal_vein_and_splenic_vein** | 0.2906 | **0.2966** | 0.0000 | **** | **0.0750** | 0.0704 | 0.0000 | **** |
| **prostate** | 0.5778 | **0.6333** | 0.0000 | **** | NA | NA | NA | NA |
| **sacrum** | 0.7485 | **0.7583** | 0.0000 | **** | 0.6883 | **0.7094** | 0.0000 | **** |
| **scapula_left** | 0.6247 | **0.7053** | 0.0000 | **** | 0.5303 | **0.6530** | 0.0000 | **** |
| **scapula_right** | 0.4408 | **0.6134** | 0.0000 | **** | 0.5442 | **0.6597** | 0.0000 | **** |
| **small_bowel** | 0.3714 | **0.3959** | 0.0000 | **** | 0.2670 | **0.2944** | 0.0000 | **** |
| **spinal_cord** | 0.7596 | **0.7679** | 0.0000 | **** | 0.7417 | **0.7477** | 0.0000 | **** |
| **spleen** | 0.6905 | **0.6994** | 0.0000 | **** | **0.6711** | 0.6618 | 0.0000 | **** |
| **stomach** | **0.5803** | 0.5762 | 0.0000 | **** | 0.4302 | **0.4374** | 0.0000 | **** |
| **urinary_bladder** | 0.6214 | **0.6373** | 0.0000 | **** | 0.6358 | **0.6485** | 0.0000 | **** |
| **vertebrae** | 0.7402 | **0.7413** | 0.0112 | * | 0.6591 | **0.6640** | 0.0000 | **** |

Supplementary Table F: Mean dice scores of the 71 masks segmented with VIBESegmentator for the female and male cohorts for the intensity-based and uniGradICON registrations. The statistical tests were done using pairwise Wilcoxon signed-rank tests with adjustment for multiple testing using Bonferroni correction. (na: not applicable, ns: not significant, *: p-value < 0.05, **: p-value < 0.01, ***: p-value < 0.001, ****: p-value < 0.0001)

| | **male** | | | | **female** | | | |
|---|---|---|---|---|---|---|---|---|
| **mask name** | **intensity based** | **uniGradICON** | **p.adj** | **significance** | **intensity based** | **uniGradICON** | **p.adj** | **significance** |
| **adrenal gland left** | **0.4685** | 0.5632 | 0.000 | **** | **0.2205** | 0.3459 | 0.000 | **** |
| **adrenal gland right** | **0.4245** | 0.5567 | 0.000 | **** | **0.2770** | 0.3372 | 0.000 | **** |
| **aorta** | **0.7590** | 0.8467 | 0.000 | **** | 0.7076 | **0.7603** | 0.000 | **** |
| **atrial appendage left** | **0.4384** | 0.6055 | 0.000 | **** | **0.4123** | 0.3926 | 0.002 | ** |
| **autochthon left** | 0.9052 | **0.9179** | 0.000 | **** | 0.8980 | **0.9056** | 0.000 | **** |
| **autochthon right** | 0.9020 | **0.9097** | 0.000 | **** | 0.8912 | **0.9074** | 0.000 | **** |
| **bone other** | 0.3753 | **0.4908** | 0.000 | **** | 0.7032 | **0.6903** | 0.000 | **** |
| **brachiocephalic trunk** | 0.4827 | **0.6492** | 0.000 | **** | 0.5923 | **0.6156** | 0.000 | **** |
| **brachiocephalic vein left** | 0.4910 | **0.6300** | 0.000 | **** | 0.5751 | **0.6352** | 0.000 | **** |
| **brachiocephalic vein right** | 0.6327 | **0.6289** | 1.000 | ns | 0.6758 | **0.7224** | 0.000 | **** |
| **clavicula left** | 0.5312 | **0.6171** | 0.000 | **** | 0.5482 | **0.5898** | 0.000 | **** |
| **clavicula right** | 0.5128 | **0.4781** | 0.000 | **** | 0.5609 | **0.6494** | 0.000 | **** |
| **common carotid artery left** | 0.4333 | **0.5517** | 0.000 | **** | 0.5320 | **0.4978** | 0.000 | **** |
| **common carotid artery right** | 0.4848 | **0.4363** | 0.000 | **** | 0.4918 | **0.5734** | 0.000 | **** |
| **costal cartilages** | 0.4748 | **0.5480** | 0.000 | **** | 0.5369 | **0.6387** | 0.000 | **** |
| **duodenum** | 0.6356 | **0.6427** | 0.000 | **** | 0.4216 | **0.4560** | 0.000 | **** |
| **esophagus** | 0.6312 | **0.7012** | 0.000 | **** | 0.6243 | **0.6724** | 0.000 | **** |
| **femur left** | 0.8718 | **0.8903** | 0.000 | **** | 0.8885 | **0.8932** | 0.630 | ns |
| **femur right** | 0.8761 | **0.8879** | 0.000 | **** | 0.8766 | **0.8885** | 0.000 | **** |
| **gallbladder** | 0.2686 | **0.3550** | 0.000 | **** | 0.2071 | **0.2107** | 1.000 | ns |
| **gluteus maximus left** | 0.9227 | **0.9297** | 0.000 | **** | 0.9288 | **0.9325** | 0.000 | **** |
| **gluteus maximus right** | 0.9350 | **0.9438** | 0.000 | **** | 0.9199 | **0.9332** | 0.000 | **** |
| **gluteus medius left** | 0.8820 | **0.8926** | 0.000 | **** | 0.8501 | **0.8827** | 0.000 | **** |
| **gluteus medius right** | 0.8893 | **0.8999** | 0.000 | **** | 0.8489 | **0.8786** | 0.000 | **** |
| **gluteus minimus left** | 0.8430 | **0.8488** | 0.000 | **** | 0.7947 | **0.8213** | 0.000 | **** |
| **gluteus minimus right** | 0.8303 | **0.8451** | 0.000 | **** | 0.8101 | **0.8235** | 0.000 | **** |
| **heart** | **0.8419** | 0.8876 | 0.000 | **** | **0.8529** | 0.8961 | 0.000 | **** |
| **hip left** | 0.8300 | **0.8625** | 0.000 | **** | 0.8093 | **0.8562** | 0.000 | **** |
| **hip right** | 0.8474 | **0.8818** | 0.000 | **** | 0.7980 | **0.8423** | 0.000 | **** |
| **humerus left** | 0.7278 | **0.8061** | 0.000 | **** | 0.7321 | **0.7408** | 1.000 | ns |
| **humerus right** | 0.5475 | **0.5791** | 0.000 | **** | 0.7313 | **0.7917** | 0.000 | **** |
| **iliac artery left** | 0.4732 | **0.5562** | 0.000 | **** | 0.5016 | **0.5802** | 0.000 | **** |
| **iliac artery right** | 0.4021 | **0.5087** | 0.000 | **** | 0.4363 | **0.5517** | 0.000 | **** |
| **iliac vena left** | 0.6088 | **0.6843** | 0.000 | **** | 0.5725 | **0.6840** | 0.000 | **** |
| **iliac vena right** | 0.5065 | **0.6117** | 0.000 | **** | 0.5049 | **0.6444** | 0.000 | **** |
| **iliopsoas left** | 0.8656 | **0.8848** | 0.000 | **** | 0.8256 | **0.8444** | 0.000 | **** |
| **iliopsoas right** | 0.8637 | **0.8808** | 0.000 | **** | 0.8201 | **0.8163** | 0.140 | ns |
| **inferior vena cava** | **0.6880** | 0.7368 | 0.000 | **** | 0.5476 | **0.5960** | 0.000 | **** |
| **inner fat** | 0.6946 | **0.7302** | 0.000 | **** | 0.5651 | **0.6303** | 0.000 | **** |
| **intestine** | 0.7230 | **0.7371** | 0.000 | **** | 0.7480 | **0.7489** | 0.000 | **** |
| **IVD** | 0.4367 | **0.6588** | 0.000 | **** | 0.3417 | **0.5405** | 0.000 | **** |
| **kidney left** | 0.8364 | **0.8531** | 0.000 | **** | **0.7234** | 0.7547 | 0.000 | **** |
| **kidney right** | 0.8317 | **0.8708** | 0.000 | **** | **0.7893** | 0.6965 | 0.000 | **** |
| **liver** | 0.8865 | **0.9103** | 0.000 | **** | 0.8432 | **0.8677** | 0.000 | **** |
| **lung lower lobe left** | 0.8652 | **0.8317** | 0.000 | **** | 0.8894 | **0.8960** | 0.000 | **** |
| **lung lower lobe right** | **0.8976** | 0.8466 | 0.000 | **** | 0.9052 | **0.8796** | 0.000 | **** |
| **lung middle lobe right** | 0.8350 | **0.8250** | 0.000 | **** | 0.8304 | **0.8284** | 1.000 | ns |
| **lung upper lobe left** | 0.9001 | **0.8963** | 0.000 | **** | 0.9070 | **0.9038** | 0.000 | **** |
| **lung upper lobe right** | 0.8959 | **0.8813** | 0.000 | **** | 0.9093 | **0.9022** | 0.000 | **** |
| **muscle** | 0.8745 | **0.8867** | 0.000 | **** | 0.8752 | **0.8769** | 1.000 | ns |
| **pancreas** | 0.6007 | **0.6206** | 0.000 | **** | **0.4185** | 0.3445 | 0.000 | **** |
| **portal vein and splenic vein** | 0.4241 | **0.5514** | 0.000 | **** | **0.4129** | 0.2840 | 0.000 | **** |
| **prostate** | 0.6367 | **0.6197** | 0.000 | **** | NA | NA | NA | NA |
| **pulmonary vein** | **0.6876** | 0.6227 | 0.000 | **** | **0.7025** | 0.6088 | 0.000 | **** |
| **sacrum** | 0.8739 | **0.8850** | 0.000 | **** | 0.7887 | **0.8374** | 0.000 | **** |
| **scapula left** | 0.7259 | **0.7844** | 0.000 | **** | 0.6719 | **0.6673** | 0.004 | ** |
| **scapula right** | 0.6176 | **0.6238** | 0.000 | **** | 0.6568 | **0.7050** | 0.000 | **** |
| **spinal channel** | 0.7190 | **0.7337** | 0.000 | **** | 0.6893 | **0.7138** | 0.000 | **** |
| **spinal cord** | 0.5971 | **0.6449** | 0.000 | **** | 0.6005 | **0.6507** | 0.000 | **** |
| **spleen** | **0.7139** | 0.7001 | 0.000 | **** | **0.5879** | 0.7005 | 0.000 | **** |
| **sternum** | 0.5649 | **0.6111** | 0.000 | **** | 0.6931 | **0.7636** | 0.000 | **** |
| **stomach** | **0.6634** | 0.6393 | 0.000 | **** | 0.4474 | **0.4958** | 0.000 | **** |
| **subclavian artery left** | 0.4611 | **0.5770** | 0.000 | **** | 0.5649 | **0.5259** | 0.000 | **** |
| **subclavian artery right** | 0.4900 | **0.5567** | 0.000 | **** | 0.5691 | **0.6519** | 0.000 | **** |
| **subcutaneous fat** | 0.8887 | **0.8795** | 0.000 | **** | 0.9220 | **0.9169** | 0.000 | **** |
| **superior vena cava** | 0.7400 | **0.7209** | 0.000 | **** | **0.7478** | 0.7825 | 0.000 | **** |
| **thyroid gland** | 0.4680 | **0.4876** | 0.000 | **** | 0.4482 | **0.5558** | 0.000 | **** |
| **trachea** | 0.7286 | **0.7283** | 0.923 | ns | 0.7237 | **0.7058** | 0.000 | **** |
| **urinary bladder** | 0.7656 | **0.6281** | 0.000 | **** | 0.8259 | **0.6975** | 0.000 | **** |
| **vertebra body** | 0.7168 | **0.7925** | 0.000 | **** | 0.6575 | **0.7165** | 0.000 | **** |
| **vertebra posterior elements** | 0.6425 | **0.6254** | 0.000 | **** | 0.5940 | **0.6145** | 0.000 | **** |

Supplementary Table G: Mean dice scores of the 71 masks segmented with VIBESegmentator for the female and male cohorts for the mask-supported and uniGradICON registrations. The statistical tests were done using pairwiseWilcoxon signed-rank tests with adjustment for multiple testing using Bonferroni correction. (na: not applicable, ns: not significant, *: p-value < 0.05, **: p-value < 0.01, ***: p-value < 0.001, ****: p-value < 0.0001)

| | male | | | | female | | | |
|---|---|---|---|---|---|---|---|---|
| mask name | uniGradICON | mask-supported | p.adj | significance | uniGradICON | mask-supported | p.adj | significance |
| adrenal gland left | 0.4685 | **0.5631** | 0.000 | **** | 0.2205 | **0.3380** | 0.000 | **** |
| adrenal gland right | 0.4245 | **0.5573** | 0.000 | **** | 0.2770 | **0.3292** | 0.000 | **** |
| aorta | 0.7590 | **0.8466** | 0.000 | **** | 0.7076 | **0.7649** | 0.000 | **** |
| atrial appendage left | 0.4384 | **0.5988** | 0.000 | **** | **0.4123** | 0.3924 | 0.002 | ** |
| autochthon left | 0.9052 | **0.9514** | 0.000 | **** | 0.8980 | **0.9397** | 0.000 | **** |
| autochthon right | 0.9020 | **0.9526** | 0.000 | **** | 0.8912 | **0.9468** | 0.000 | **** |
| bone other | 0.3753 | **0.6070** | 0.000 | **** | 0.7032 | **0.8310** | 0.000 | **** |
| brachiocephalic trunk | 0.4827 | **0.6817** | 0.000 | **** | 0.5923 | **0.6278** | 0.000 | **** |
| brachiocephalic vein left | 0.4910 | **0.6774** | 0.000 | **** | 0.5751 | **0.6653** | 0.000 | **** |
| brachiocephalic vein right | 0.6327 | **0.7202** | 0.000 | **** | 0.6758 | **0.7340** | 0.000 | **** |
| clavicula left | 0.5312 | **0.8372** | 0.000 | **** | 0.5482 | **0.8031** | 0.000 | **** |
| clavicula right | 0.5128 | **0.8131** | 0.000 | **** | 0.5609 | **0.8541** | 0.000 | **** |
| common carotid artery left | 0.4333 | **0.7214** | 0.000 | **** | 0.5320 | **0.6460** | 0.000 | **** |
| common carotid artery right | 0.4848 | **0.7618** | 0.000 | **** | 0.4918 | **0.7743** | 0.000 | **** |
| costal cartilages | 0.4748 | **0.7067** | 0.000 | **** | 0.5369 | **0.7535** | 0.000 | **** |
| duodenum | 0.6356 | **0.6610** | 0.000 | **** | 0.4216 | **0.4641** | 0.000 | **** |
| esophagus | 0.6312 | **0.7396** | 0.000 | **** | 0.6243 | **0.6969** | 0.000 | **** |
| femur left | 0.8718 | **0.9281** | 0.000 | **** | 0.8885 | **0.9656** | 0.000 | **** |
| femur right | 0.8761 | **0.9042** | 0.000 | **** | 0.8766 | **0.9602** | 0.000 | **** |
| gallbladder | 0.2686 | **0.3624** | 0.000 | **** | 0.2071 | **0.2199** | 0.000 | *** |
| gluteus maximus left | 0.9227 | **0.9651** | 0.000 | **** | 0.9288 | **0.9704** | 0.000 | **** |
| gluteus maximus right | 0.9350 | **0.9662** | 0.000 | **** | 0.9199 | **0.9717** | 0.000 | **** |
| gluteus medius left | 0.8820 | **0.9197** | 0.000 | **** | 0.8501 | **0.9155** | 0.000 | **** |
| gluteus medius right | 0.8893 | **0.9184** | 0.000 | **** | 0.8489 | **0.9124** | 0.000 | **** |
| gluteus minimus left | 0.8430 | **0.8678** | 0.000 | **** | 0.7947 | **0.8512** | 0.000 | **** |
| gluteus minimus right | 0.8303 | **0.8589** | 0.000 | **** | 0.8101 | **0.8569** | 0.000 | **** |
| heart | 0.8419 | **0.8876** | 0.000 | **** | 0.8529 | **0.8947** | 0.000 | **** |
| hip left | 0.8300 | **0.8991** | 0.000 | **** | 0.8093 | **0.9053** | 0.000 | **** |
| hip right | 0.8474 | **0.9015** | 0.000 | **** | 0.7980 | **0.8975** | 0.000 | **** |
| humerus left | 0.7278 | **0.9283** | 0.000 | **** | 0.7321 | **0.9209** | 0.000 | **** |
| humerus right | 0.5475 | **0.8026** | 0.000 | **** | 0.7313 | **0.9265** | 0.000 | **** |
| iliac artery left | 0.4732 | **0.5825** | 0.000 | **** | 0.5016 | **0.5986** | 0.000 | **** |
| iliac artery right | 0.4021 | **0.5375** | 0.000 | **** | 0.4363 | **0.5890** | 0.000 | **** |
| **iliac vena left** | 0.6088 | **0.6999** | 0.000 | **** | 0.5725 | **0.7021** | 0.000 | **** |
| **iliac vena right** | 0.5065 | **0.6239** | 0.000 | **** | 0.5049 | **0.6669** | 0.000 | **** |
| **iliopsoas left** | 0.8656 | **0.8999** | 0.000 | **** | 0.8256 | **0.8637** | 0.000 | **** |
| **iliopsoas right** | 0.8637 | **0.8916** | 0.000 | **** | 0.8201 | **0.8403** | 0.000 | **** |
| **inferior vena cava** | 0.6880 | **0.7339** | 0.000 | **** | 0.5476 | **0.5991** | 0.000 | **** |
| **inner fat** | 0.6946 | **0.7684** | 0.000 | **** | 0.5651 | **0.6910** | 0.000 | **** |
| **intestine** | 0.7230 | **0.7818** | 0.000 | **** | 0.7480 | **0.7963** | 0.000 | **** |
| **IVD** | 0.4367 | **0.6615** | 0.000 | **** | 0.3417 | **0.5463** | 0.000 | **** |
| **kidney left** | 0.8364 | **0.8636** | 0.000 | **** | 0.7234 | **0.7452** | 0.000 | **** |
| **kidney right** | 0.8317 | **0.8753** | 0.000 | **** | **0.7893** | 0.6969 | 0.000 | **** |
| **liver** | 0.8865 | **0.9143** | 0.000 | **** | 0.8432 | **0.8722** | 0.000 | **** |
| **lung lower lobe left** | **0.8652** | 0.8343 | 0.000 | **** | 0.8894 | **0.9018** | 0.000 | **** |
| **lung lower lobe right** | **0.8976** | 0.8468 | 0.000 | **** | **0.9052** | 0.8858 | 0.000 | **** |
| **lung middle lobe right** | **0.8350** | 0.8313 | 0.000 | **** | 0.8304 | **0.8369** | 0.000 | **** |
| **lung upper lobe left** | 0.9001 | **0.9092** | 0.000 | **** | 0.9070 | **0.9171** | 0.000 | **** |
| **lung upper lobe right** | **0.8959** | 0.8924 | 0.000 | **** | 0.9093 | **0.9121** | 0.000 | **** |
| **muscle** | 0.8745 | **0.9647** | 0.000 | **** | 0.8752 | **0.9658** | 0.000 | **** |
| **pancreas** | 0.6007 | **0.6298** | 0.000 | **** | **0.4185** | 0.3483 | 0.000 | **** |
| **portal vein and splenic vein** | 0.4241 | **0.5564** | 0.000 | **** | **0.4129** | 0.2775 | 0.000 | **** |
| **prostate** | 0.6367 | **0.6759** | 0.000 | **** | NA | NA | NA | NA |
| **pulmonary vein** | **0.6876** | 0.6217 | 0.000 | **** | **0.7025** | 0.6068 | 0.000 | **** |
| **sacrum** | 0.8739 | **0.8963** | 0.000 | **** | 0.7887 | **0.8592** | 0.000 | **** |
| **scapula left** | 0.7259 | **0.9411** | 0.000 | **** | 0.6719 | **0.8947** | 0.000 | **** |
| **scapula right** | 0.6176 | **0.8781** | 0.000 | **** | 0.6568 | **0.9077** | 0.000 | **** |
| **spinal channel** | 0.7190 | **0.7500** | 0.000 | **** | 0.6893 | **0.7259** | 0.000 | **** |
| **spinal cord** | 0.5971 | **0.6762** | 0.000 | **** | 0.6005 | **0.6697** | 0.000 | **** |
| **spleen** | **0.7139** | 0.7064 | 0.000 | **** | 0.5879 | **0.6985** | 0.000 | **** |
| **sternum** | 0.5649 | **0.7286** | 0.000 | **** | 0.6931 | **0.8474** | 0.000 | **** |
| **stomach** | **0.6634** | 0.6361 | 0.000 | **** | 0.4474 | **0.5070** | 0.000 | **** |
| **subclavian artery left** | 0.4611 | **0.7504** | 0.000 | **** | 0.5649 | **0.6708** | 0.000 | **** |
| **subclavian artery right** | 0.4900 | **0.7990** | 0.000 | **** | 0.5691 | **0.7993** | 0.000 | **** |
| **subcutaneous fat** | 0.8887 | **0.9628** | 0.000 | **** | 0.9220 | **0.9756** | 0.000 | **** |
| **superior vena cava** | **0.7400** | 0.7281 | 0.000 | **** | 0.7478 | **0.7826** | 0.000 | **** |
| **thyroid gland** | 0.4680 | **0.8294** | 0.000 | **** | 0.4482 | **0.8086** | 0.000 | **** |
| **trachea** | 0.7286 | **0.7985** | 0.000 | **** | 0.7237 | **0.7455** | 0.000 | **** |
| **urinary bladder** | **0.7656** | 0.6433 | 0.000 | **** | **0.8259** | 0.7086 | 0.000 | **** |
| **vertebra body** | 0.7168 | **0.8001** | 0.000 | **** | 0.6575 | **0.7252** | 0.000 | **** |
| **vertebra posterior elements** | 0.6425 | **0.6495** | 0.000 | **** | 0.5940 | **0.6329** | 0.000 | **** |

Supplementary Table H: Mean dice scores of the 40 masks segmented with MRSegmentator for the female and male cohorts for the mask-supported and uniGradICON registrations. The statistical tests were done using pairwise Wilcoxon signed-rank tests with adjustment for multiple testing using Bonferroni correction. (na: not applicable, ns: not significant, *: p-value < 0.05, **: p-value < 0.01, ***: p-value < 0.001, ****: p-value < 0.0001)

| | male | | | | female | | | |
|---|---|---|---|---|---|---|---|---|
| mask name | uniGradICON | mask supported | p.adj | significance | uniGradICON | mask supported | p.adj | significance |
| aorta | 0.7261 | **0.8476** | 0.0000 | **** | 0.6693 | **0.7701** | 0.0000 | **** |
| colon | 0.4807 | **0.5122** | 0.0000 | **** | **0.4404** | 0.4312 | 0.0000 | **** |
| duodenum | **0.5719** | 0.5616 | 0.0149 | * | 0.3257 | **0.4316** | 0.0000 | **** |
| esophagus | 0.5351 | **0.6550** | 0.0000 | **** | 0.5194 | **0.5689** | 0.0000 | **** |
| gallbladder | 0.2358 | **0.2485** | 0.0050 | ** | **0.1671** | 0.1198 | 0.0000 | **** |
| heart | 0.8271 | **0.8880** | 0.0000 | **** | 0.8424 | **0.8835** | 0.0000 | **** |
| inferior_vena_cava | 0.6118 | **0.6760** | 0.0000 | **** | 0.3798 | **0.4965** | 0.0000 | **** |
| left_adrenal_gland | 0.3692 | **0.5170** | 0.0000 | **** | NA | NA | NA | NA |
| left_autochthonous_muscle | 0.9055 | **0.9102** | 0.0000 | **** | 0.8944 | **0.9017** | 0.0000 | **** |
| left_femur | 0.8516 | **0.8838** | 0.0000 | **** | 0.8775 | **0.9151** | 0.0000 | **** |
| left_gluteus_maximus | 0.9144 | **0.9234** | 0.0000 | **** | 0.9162 | **0.9247** | 0.0000 | **** |
| left_gluteus_medius | 0.8795 | **0.8878** | 0.0000 | **** | 0.8562 | **0.8910** | 0.0000 | **** |
| left_gluteus_minimus | 0.7912 | **0.8054** | 0.0000 | **** | 0.7315 | **0.7720** | 0.0000 | **** |
| left_hip | 0.8061 | **0.8676** | 0.0000 | **** | 0.7862 | **0.8706** | 0.0000 | **** |
| left_iliac_artery | 0.3251 | **0.4966** | 0.0000 | **** | 0.3305 | **0.4952** | 0.0000 | **** |
| left_iliac_vena | 0.5484 | **0.6530** | 0.0000 | **** | 0.5197 | **0.6830** | 0.0000 | **** |
| left_iliopsoas_muscle | 0.8550 | **0.8873** | 0.0000 | **** | 0.8144 | **0.8490** | 0.0000 | **** |
| left_kidney | 0.8540 | **0.8672** | 0.0000 | **** | 0.7224 | **0.7473** | 0.0000 | **** |
| left_lung | **0.9446** | 0.9097 | 0.0000 | **** | **0.9512** | 0.9390 | 0.0000 | **** |
| liver | 0.8937 | **0.9091** | 0.0000 | **** | 0.8533 | **0.8673** | 0.0000 | **** |
| pancreas | 0.5126 | **0.5279** | 0.0000 | **** | 0.2786 | **0.3001** | 0.0000 | **** |
| portal_vein_and_splenic_vein | 0.2400 | **0.3713** | 0.0000 | **** | **0.1907** | 0.1479 | 0.0000 | **** |
| right_adrenal_gland | 0.4123 | **0.4989** | 0.0000 | **** | 0.1366 | **0.2147** | 0.0000 | **** |
| right_autochthonous_muscle | 0.8969 | **0.8982** | 0.0000 | **** | **0.8927** | 0.8869 | 0.0000 | **** |
| right_femur | 0.8639 | **0.8798** | 0.0000 | **** | 0.8727 | **0.9110** | 0.0000 | **** |
| right_gluteus_maximus | 0.9294 | **0.9360** | 0.0000 | **** | 0.9084 | **0.9260** | 0.0000 | **** |
| right_gluteus_medius | 0.8824 | **0.9046** | 0.0000 | **** | 0.8523 | **0.8895** | 0.0000 | **** |
| right_gluteus_minimus | 0.8001 | **0.8237** | 0.0000 | **** | 0.7661 | **0.7881** | 0.0000 | **** |
| right_hip | 0.8210 | **0.8763** | 0.0000 | **** | 0.7684 | **0.8613** | 0.0000 | **** |
| right_iliac_artery | 0.2461 | **0.4013** | 0.0000 | **** | 0.2639 | **0.4856** | 0.0000 | **** |
| right_iliac_vena | 0.3766 | **0.5453** | 0.0000 | **** | 0.3789 | **0.5931** | 0.0000 | **** |
| right_iliopsoas_muscle | 0.8571 | **0.8819** | 0.0000 | **** | 0.8061 | **0.8372** | 0.0000 | **** |
| right_kidney | 0.8447 | **0.8824** | 0.0000 | **** | **0.7831** | 0.6877 | 0.0000 | **** |
| **right_lung** | **0.9570** | 0.9223 | 0.0000 | **** | **0.9624** | 0.9406 | 0.0000 | **** |
| **sacrum** | 0.8160 | **0.8482** | 0.0000 | **** | 0.7061 | **0.8101** | 0.0000 | **** |
| **small_bowel** | 0.5080 | **0.5639** | 0.0000 | **** | 0.5015 | **0.5325** | 0.0000 | **** |
| **spine** | 0.6959 | **0.7831** | 0.0000 | **** | 0.6142 | **0.6946** | 0.0000 | **** |
| **spleen** | 0.7167 | **0.7186** | 0.0000 | **** | 0.5924 | **0.7002** | 0.0000 | **** |
| **stomach** | **0.6591** | 0.5982 | 0.0000 | **** | 0.4224 | **0.4616** | 0.0000 | **** |
| **urinary_bladder** | **0.7567** | 0.6291 | 0.0000 | **** | **0.8065** | 0.6687 | 0.0000 | **** |

Supplementary Table I: Mean dice scores of the 50 masks segmented with TotalSegmentator for the female and male cohorts for the mask-supported and uniGradICON registrations. The statistical tests were done using pairwise Wilcoxon signed-rank tests with adjustment for multiple testing using Bonferroni correction. (na: not applicable, ns: not significant, *: p-value < 0.05, **: p-value < 0.01, ***: p-value < 0.001, ****: p-value < 0.0001)

| | male | | | | female | | | |
|---|---|---|---|---|---|---|---|---|
| **mask name** | **uniGradICON** | **mask supported** | **p.adj** | **significance** | **uniGradICON** | **mask supported** | **p.adj** | **significance** |
| **adrenal_gland_left** | 0.3529 | **0.4339** | 0.0000 | **** | 0.0565 | **0.1384** | 0.0000 | **** |
| **adrenal_gland_right** | 0.3992 | **0.4516** | 0.0000 | **** | 0.1610 | **0.2034** | 0.0000 | **** |
| **aorta** | 0.6532 | **0.8005** | 0.0000 | **** | 0.5924 | **0.7062** | 0.0000 | **** |
| **autochthon_left** | 0.9013 | **0.9068** | 0.0000 | **** | 0.8856 | **0.8929** | 0.0000 | **** |
| **autochthon_right** | 0.8912 | **0.8994** | 0.0000 | **** | **0.8785** | 0.8762 | 0.8640 | ns |
| **clavicula_left** | 0.3037 | **0.5717** | 0.0000 | **** | 0.2898 | **0.5150** | 0.0000 | **** |
| **clavicula_right** | 0.2494 | **0.5050** | 0.0000 | **** | 0.2707 | **0.5156** | 0.0000 | **** |
| **colon** | 0.4580 | **0.4759** | 0.0000 | **** | **0.4092** | 0.3747 | 0.0000 | **** |
| **duodenum** | **0.5184** | 0.4763 | 0.0000 | **** | 0.1721 | **0.2574** | 0.0000 | **** |
| **esophagus** | 0.5286 | **0.6156** | 0.0000 | **** | 0.5081 | **0.5442** | 0.0000 | **** |
| **femur_left** | 0.8338 | **0.8722** | 0.0000 | **** | 0.8570 | **0.8972** | 0.0000 | **** |
| **femur_right** | 0.8550 | **0.8758** | 0.0000 | **** | 0.8633 | **0.9038** | 0.0000 | **** |
| **gallbladder** | 0.2208 | **0.2269** | 0.6860 | ns | **0.1804** | 0.1199 | 0.0000 | **** |
| **gluteus_maximus_left** | 0.9064 | **0.9145** | 0.0000 | **** | 0.9117 | **0.9189** | 0.0000 | **** |
| **gluteus_maximus_right** | 0.9096 | **0.9178** | 0.0000 | **** | 0.8892 | **0.9069** | 0.0000 | **** |
| **gluteus_medius_left** | 0.8192 | **0.8269** | 0.0000 | **** | 0.7301 | **0.7589** | 0.0000 | **** |
| **gluteus_medius_right** | 0.8254 | **0.8455** | 0.0000 | **** | 0.7025 | **0.7458** | 0.0000 | **** |
| **gluteus_minimus_left** | 0.7464 | **0.7703** | 0.0000 | **** | 0.6695 | **0.7179** | 0.0000 | **** |
| **gluteus_minimus_right** | 0.7467 | **0.7644** | 0.0000 | **** | 0.6963 | **0.7219** | 0.0000 | **** |
| **heart** | 0.8237 | **0.8678** | 0.0000 | **** | 0.8144 | **0.8632** | 0.0000 | **** |
| **hip_left** | 0.7432 | **0.8251** | 0.0000 | **** | 0.7176 | **0.8027** | 0.0000 | **** |
| **hip_right** | 0.7546 | **0.8312** | 0.0000 | **** | 0.6828 | **0.7971** | 0.0000 | **** |
| **humerus_left** | 0.7321 | **0.8263** | 0.0000 | **** | 0.7072 | **0.8010** | 0.0000 | **** |
| **humerus_right** | 0.5148 | **0.6805** | 0.0000 | **** | 0.6983 | **0.8181** | 0.0000 | **** |
| **iliac_artery_left** | 0.2806 | **0.4032** | 0.0000 | **** | 0.2461 | **0.3290** | 0.0000 | **** |
| **iliac_artery_right** | 0.2182 | **0.3359** | 0.0000 | **** | 0.1958 | **0.3546** | 0.0000 | **** |
| **iliac_vena_left** | 0.4895 | **0.5562** | 0.0000 | **** | 0.4438 | **0.5605** | 0.0000 | **** |
| **iliac_vena_right** | 0.3448 | **0.5024** | 0.0000 | **** | 0.3368 | **0.4995** | 0.0000 | **** |
| **iliopsoas_left** | 0.8274 | **0.8563** | 0.0000 | **** | 0.7907 | **0.8243** | 0.0000 | **** |
| **iliopsoas_right** | 0.8335 | **0.8666** | 0.0000 | **** | 0.7898 | **0.8114** | 0.0000 | **** |
| **inferior_vena_cava** | 0.5370 | **0.5723** | 0.0000 | **** | 0.1976 | **0.2391** | 0.0000 | **** |
| **intervertebral_discs** | 0.4783 | **0.6174** | 0.0000 | **** | 0.3790 | **0.5059** | 0.0000 | **** |
| **kidney_left** | 0.8002 | **0.8224** | 0.0000 | **** | 0.6530 | **0.6746** | 0.0000 | **** |

| **kidney_right** | 0.8103 | **0.8291** | 0.0000 | **** | **0.7247** | 0.6057 | 0.0000 | **** |
|---|---|---|---|---|---|---|---|---|
| **liver** | 0.8818 | **0.8918** | 0.0000 | **** | **0.8511** | 0.8498 | 0.0010 | *** |
| **lung_left** | **0.9451** | 0.9086 | 0.0000 | **** | **0.9497** | 0.9362 | 0.0000 | **** |
| **lung_right** | **0.9571** | 0.9229 | 0.0000 | **** | **0.9614** | 0.9445 | 0.0000 | **** |
| **pancreas** | 0.3641 | **0.3653** | 0.2940 | ns | **0.0657** | 0.0588 | 0.0000 | **** |
| **portal_vein_and_splenic_vein** | 0.1796 | **0.2966** | 0.0000 | **** | NA | NA | NA | NA |
| **prostate** | 0.5696 | **0.6333** | 0.0000 | **** | **0.1449** | 0.0704 | 0.0000 | **** |
| **sacrum** | 0.7532 | **0.7583** | 0.0000 | **** | 0.6195 | **0.7094** | 0.0000 | **** |
| **scapula_left** | 0.5381 | **0.7053** | 0.0000 | **** | 0.4940 | **0.6530** | 0.0000 | **** |
| **scapula_right** | 0.4313 | **0.6134** | 0.0000 | **** | 0.4907 | **0.6597** | 0.0000 | **** |
| **small_bowel** | 0.3594 | **0.3959** | 0.0000 | **** | **0.2956** | 0.2944 | 1.0000 | ns |
| **spinal_cord** | 0.7645 | **0.7679** | 0.0022 | ** | 0.7223 | **0.7477** | 0.0000 | **** |
| **spleen** | 0.6992 | **0.6994** | 0.0000 | **** | 0.5381 | **0.6618** | 0.0000 | **** |
| **stomach** | **0.6257** | 0.5762 | 0.0000 | **** | 0.4089 | **0.4374** | 0.0000 | **** |
| **urinary_bladder** | **0.7718** | 0.6373 | 0.0000 | **** | **0.7901** | 0.6485 | 0.0000 | **** |
| **vertebrae** | 0.6757 | **0.7413** | 0.0000 | **** | 0.6196 | **0.6640** | 0.0000 | **** |

Supplementary Table J: Mean dice scores of the 71 masks segmented with VIBESegmentator for the female and male cohorts for the intensity-based and MIRTK registrations. The statistical tests were done using pairwise Wilcoxon signed-rank tests with adjustment for multiple testing using Bonferroni correction. (na: not applicable, ns: not significant, *: p-value < 0.05, **: p-value < 0.01, ***: p-value < 0.001, ****: p-value < 0.0001)

| | male | | | | female | | | |
|---|---|---|---|---|---|---|---|---|
| mask name | intensity based | MIRTK | p.adj | significance | intensity based | MIRTK | p.adj | significance |
| adrenal gland left | **0.4022** | 0.5632 | 0.000 | **** | **0.1718** | 0.3459 | 0.000 | **** |
| adrenal gland right | **0.3399** | 0.5567 | 0.000 | **** | **0.1639** | 0.3372 | 0.000 | **** |
| aorta | **0.7279** | 0.8467 | 0.000 | **** | 0.6393 | **0.7603** | 0.000 | **** |
| atrial appendage left | **0.4830** | 0.6055 | 0.000 | **** | **0.4388** | 0.3926 | 0.000 | **** |
| autochthon left | 0.8852 | **0.9179** | 0.000 | **** | 0.8349 | **0.9056** | 0.000 | **** |
| autochthon right | 0.8734 | **0.9097** | 0.000 | **** | 0.8260 | **0.9074** | 0.000 | **** |
| bone other | 0.5647 | **0.4908** | 0.000 | **** | 0.7167 | **0.6903** | 0.000 | **** |
| brachiocephalic trunk | 0.5509 | **0.6492** | 0.000 | **** | 0.5007 | **0.6156** | 0.000 | **** |
| brachiocephalic vein left | 0.5535 | **0.6300** | 0.000 | **** | 0.5058 | **0.6352** | 0.000 | **** |
| brachiocephalic vein right | 0.6335 | **0.6289** | 1.000 | ns | 0.5708 | **0.7224** | 0.000 | **** |
| clavicula left | 0.6674 | **0.6171** | 0.000 | **** | 0.6149 | **0.5898** | 0.020 | * |
| clavicula right | 0.6622 | **0.4781** | 0.000 | **** | 0.6508 | **0.6494** | 1.000 | ns |
| common carotid artery left | 0.5379 | **0.5517** | 0.001 | ** | 0.4609 | **0.4978** | 0.000 | **** |
| common carotid artery right | 0.4742 | **0.4363** | 0.000 | **** | 0.4577 | **0.5734** | 0.000 | **** |
| costal cartilages | 0.5471 | **0.5480** | 1.000 | ns | 0.3885 | **0.6387** | 0.000 | **** |
| duodenum | 0.4967 | **0.6427** | 0.000 | **** | 0.3766 | **0.4560** | 0.000 | **** |
| esophagus | 0.5826 | **0.7012** | 0.000 | **** | 0.5521 | **0.6724** | 0.000 | **** |
| femur left | 0.8724 | **0.8903** | 0.000 | **** | 0.8645 | **0.8932** | 0.000 | **** |
| femur right | 0.8675 | **0.8879** | 0.000 | **** | 0.8621 | **0.8885** | 0.000 | **** |
| gallbladder | 0.2044 | **0.3550** | 0.000 | **** | 0.1252 | **0.2107** | 0.000 | **** |
| gluteus maximus left | 0.8617 | **0.9297** | 0.000 | **** | 0.8657 | **0.9325** | 0.000 | **** |
| gluteus maximus right | 0.8881 | **0.9438** | 0.000 | **** | 0.8246 | **0.9332** | 0.000 | **** |
| gluteus medius left | 0.8559 | **0.8926** | 0.000 | **** | 0.8061 | **0.8827** | 0.000 | **** |
| gluteus medius right | 0.8510 | **0.8999** | 0.000 | **** | 0.7704 | **0.8786** | 0.000 | **** |
| gluteus minimus left | 0.8220 | **0.8488** | 0.000 | **** | 0.7573 | **0.8213** | 0.000 | **** |
| gluteus minimus right | 0.8075 | **0.8451** | 0.000 | **** | 0.7589 | **0.8235** | 0.000 | **** |
| heart | **0.8185** | 0.8876 | 0.000 | **** | **0.8296** | 0.8961 | 0.000 | **** |
| hip left | 0.8186 | **0.8625** | 0.000 | **** | 0.7588 | **0.8562** | 0.000 | **** |
| hip right | 0.8336 | **0.8818** | 0.000 | **** | 0.7258 | **0.8423** | 0.000 | **** |
| humerus left | 0.7869 | **0.8061** | 0.000 | **** | 0.7603 | **0.7408** | 0.000 | **** |
| humerus right | 0.7508 | **0.5791** | 0.000 | **** | 0.6993 | **0.7917** | 0.000 | **** |
| iliac artery left | 0.3436 | **0.5562** | 0.000 | **** | 0.4420 | **0.5802** | 0.000 | **** |
| **iliac artery right** | 0.3548 | **0.5087** | 0.000 | **** | 0.4501 | **0.5517** | 0.000 | **** |
| **iliac vena left** | 0.4519 | **0.6843** | 0.000 | **** | 0.4994 | **0.6840** | 0.000 | **** |
| **iliac vena right** | 0.4356 | **0.6117** | 0.000 | **** | 0.5411 | **0.6444** | 0.000 | **** |
| **iliopsoas left** | 0.8189 | **0.8848** | 0.000 | **** | 0.7518 | **0.8444** | 0.000 | **** |
| **iliopsoas right** | 0.8053 | **0.8808** | 0.000 | **** | 0.7399 | **0.8163** | 0.000 | **** |
| **inferior vena cava** | **0.6087** | 0.7368 | 0.000 | **** | 0.4614 | **0.5960** | 0.000 | **** |
| **inner fat** | 0.5350 | **0.7302** | 0.000 | **** | 0.4366 | **0.6303** | 0.000 | **** |
| **intestine** | 0.6201 | **0.7371** | 0.000 | **** | 0.7069 | **0.7489** | 0.000 | **** |
| **IVD** | 0.6210 | **0.6588** | 0.000 | **** | 0.4118 | **0.5405** | 0.000 | **** |
| **kidney left** | 0.6541 | **0.8531** | 0.000 | **** | **0.5605** | 0.7547 | 0.000 | **** |
| **kidney right** | 0.6221 | **0.8708** | 0.000 | **** | **0.5938** | 0.6965 | 0.000 | **** |
| **liver** | 0.8242 | **0.9103** | 0.000 | **** | 0.7980 | **0.8677** | 0.000 | **** |
| **lung lower lobe left** | 0.7725 | **0.8317** | 0.000 | **** | 0.8056 | **0.8960** | 0.000 | **** |
| **lung lower lobe right** | **0.8127** | 0.8466 | 0.000 | **** | 0.8528 | **0.8796** | 0.000 | **** |
| **lung middle lobe right** | 0.7459 | **0.8250** | 0.000 | **** | 0.7381 | **0.8284** | 0.000 | **** |
| **lung upper lobe left** | 0.8706 | **0.8963** | 0.000 | **** | 0.8563 | **0.9038** | 0.000 | **** |
| **lung upper lobe right** | 0.8487 | **0.8813** | 0.000 | **** | 0.8538 | **0.9022** | 0.000 | **** |
| **muscle** | 0.8231 | **0.8867** | 0.000 | **** | 0.7978 | **0.8769** | 0.000 | **** |
| **pancreas** | 0.5243 | **0.6206** | 0.000 | **** | **0.3208** | 0.3445 | 0.000 | *** |
| **portal vein and splenic vein** | 0.3939 | **0.5514** | 0.000 | **** | **0.2833** | 0.2840 | 1.000 | ns |
| **prostate** | 0.6175 | **0.6197** | 1.000 | ns | NA | NA | NA | NA |
| **pulmonary vein** | **0.5851** | 0.6227 | 0.000 | **** | 0.5908 | 0.6088 | 0.000 | **** |
| **sacrum** | 0.8403 | **0.8850** | 0.000 | **** | **0.8040** | 0.8374 | 0.000 | **** |
| **scapula left** | 0.7722 | **0.7844** | 0.000 | **** | 0.7003 | **0.6673** | 0.000 | **** |
| **scapula right** | 0.7284 | **0.6238** | 0.000 | **** | 0.6921 | **0.7050** | 0.000 | **** |
| **spinal channel** | 0.6909 | **0.7337** | 0.000 | **** | 0.4940 | **0.7138** | 0.000 | **** |
| **spinal cord** | 0.5622 | **0.6449** | 0.000 | **** | 0.4678 | **0.6507** | 0.000 | **** |
| **spleen** | **0.4957** | 0.7001 | 0.000 | **** | 0.5162 | **0.7005** | 0.000 | **** |
| **sternum** | 0.7157 | **0.6111** | 0.000 | **** | **0.6461** | 0.7636 | 0.000 | **** |
| **stomach** | **0.5026** | 0.6393 | 0.000 | **** | 0.4130 | **0.4958** | 0.000 | **** |
| **subclavian artery left** | 0.5667 | **0.5770** | 0.497 | ns | 0.5318 | **0.5259** | 1.000 | ns |
| **subclavian artery right** | 0.5296 | **0.5567** | 0.000 | **** | 0.5403 | **0.6519** | 0.000 | **** |
| **subcutaneous fat** | 0.7751 | **0.8795** | 0.000 | **** | 0.8240 | **0.9169** | 0.000 | **** |
| **superior vena cava** | 0.6578 | **0.7209** | 0.000 | **** | 0.6016 | **0.7825** | 0.000 | **** |
| **thyroid gland** | 0.5087 | **0.4876** | 1.000 | ns | **0.4455** | 0.5558 | 0.000 | **** |
| **trachea** | 0.6730 | **0.7283** | 0.000 | **** | 0.5838 | **0.7058** | 0.000 | **** |
| **urinary bladder** | 0.5699 | **0.6281** | 0.000 | **** | 0.6191 | **0.6975** | 0.000 | **** |
| **vertebra body** | 0.7601 | **0.7925** | 0.000 | **** | 0.6032 | **0.7165** | 0.000 | **** |
| **vertebra posterior elements** | 0.6488 | **0.6254** | 0.000 | **** | 0.4961 | **0.6145** | 0.000 | **** |

Supplementary Table K: Mean dice scores of the 71 masks segmented with VIBESegmentator for the female and male cohorts for the mask-supported and MIRTK registrations. The statistical tests were done using pairwise Wilcoxon signed-rank tests with adjustment for multiple testing using Bonferroni correction. (na: not applicable, ns: not significant, *: p-value < 0.05, **: p-value < 0.01, ***: p-value < 0.001, ****: p-value < 0.0001)

| | male | | | | female | | | |
|---|---|---|---|---|---|---|---|---|
| mask name | MIRTK | mask-supported | p.adj | significance | MIRTK | mask-supported | p.adj | significance |
| adrenal gland left | 0.4022 | **0.5631** | 0.000 | **** | 0.1718 | **0.3380** | 0.000 | **** |
| adrenal gland right | 0.3399 | **0.5573** | 0.000 | **** | 0.1639 | **0.3292** | 0.000 | **** |
| aorta | 0.7279 | **0.8466** | 0.000 | **** | 0.6393 | **0.7649** | 0.000 | **** |
| atrial appendage left | 0.4830 | **0.5988** | 0.000 | **** | **0.4388** | 0.3924 | 0.000 | **** |
| autochthon left | 0.8852 | **0.9514** | 0.000 | **** | 0.8349 | **0.9397** | 0.000 | **** |
| autochthon right | 0.8734 | **0.9526** | 0.000 | **** | 0.8260 | **0.9468** | 0.000 | **** |
| bone other | 0.5647 | **0.6070** | 0.000 | **** | 0.7167 | **0.8310** | 0.000 | **** |
| brachiocephalic trunk | 0.5509 | **0.6817** | 0.000 | **** | 0.5007 | **0.6278** | 0.000 | **** |
| brachiocephalic vein left | 0.5535 | **0.6774** | 0.000 | **** | 0.5058 | **0.6653** | 0.000 | **** |
| brachiocephalic vein right | 0.6335 | **0.7202** | 0.000 | **** | 0.5708 | **0.7340** | 0.000 | **** |
| clavicula left | 0.6674 | **0.8372** | 0.000 | **** | 0.6149 | **0.8031** | 0.000 | **** |
| clavicula right | 0.6622 | **0.8131** | 0.000 | **** | 0.6508 | **0.8541** | 0.000 | **** |
| common carotid artery left | 0.5379 | **0.7214** | 0.000 | **** | 0.4609 | **0.6460** | 0.000 | **** |
| common carotid artery right | 0.4742 | **0.7618** | 0.000 | **** | 0.4577 | **0.7743** | 0.000 | **** |
| costal cartilages | 0.5471 | **0.7067** | 0.000 | **** | 0.3885 | **0.7535** | 0.000 | **** |
| duodenum | 0.4967 | **0.6610** | 0.000 | **** | 0.3766 | **0.4641** | 0.000 | **** |
| esophagus | 0.5826 | **0.7396** | 0.000 | **** | 0.5521 | **0.6969** | 0.000 | **** |
| femur left | 0.8724 | **0.9281** | 0.000 | **** | 0.8645 | **0.9656** | 0.000 | **** |
| femur right | 0.8675 | **0.9042** | 0.000 | **** | 0.8621 | **0.9602** | 0.000 | **** |
| gallbladder | 0.2044 | **0.3624** | 0.000 | **** | 0.1252 | **0.2199** | 0.000 | **** |
| gluteus maximus left | 0.8617 | **0.9651** | 0.000 | **** | 0.8657 | **0.9704** | 0.000 | **** |
| gluteus maximus right | 0.8881 | **0.9662** | 0.000 | **** | 0.8246 | **0.9717** | 0.000 | **** |
| gluteus medius left | 0.8559 | **0.9197** | 0.000 | **** | 0.8061 | **0.9155** | 0.000 | **** |
| gluteus medius right | 0.8510 | **0.9184** | 0.000 | **** | 0.7704 | **0.9124** | 0.000 | **** |
| gluteus minimus left | 0.8220 | **0.8678** | 0.000 | **** | 0.7573 | **0.8512** | 0.000 | **** |
| gluteus minimus right | 0.8075 | **0.8589** | 0.000 | **** | 0.7589 | **0.8569** | 0.000 | **** |
| heart | 0.8185 | **0.8876** | 0.000 | **** | 0.8296 | **0.8947** | 0.000 | **** |
| hip left | 0.8186 | **0.8991** | 0.000 | **** | 0.7588 | **0.9053** | 0.000 | **** |
| hip right | 0.8336 | **0.9015** | 0.000 | **** | 0.7258 | **0.8975** | 0.000 | **** |
| humerus left | 0.7869 | **0.9283** | 0.000 | **** | 0.7603 | **0.9209** | 0.000 | **** |
| humerus right | 0.7508 | **0.8026** | 0.000 | **** | 0.6993 | **0.9265** | 0.000 | **** |
| iliac artery left | 0.3436 | **0.5825** | 0.000 | **** | 0.4420 | **0.5986** | 0.000 | **** |
| iliac artery right | 0.3548 | **0.5375** | 0.000 | **** | 0.4501 | **0.5890** | 0.000 | **** |

| **iliac vena left** | 0.4519 | **0.6999** | 0.000 | **** | 0.4994 | **0.7021** | 0.000 | **** |
|---|---|---|---|---|---|---|---|---|
| **iliac vena right** | 0.4356 | **0.6239** | 0.000 | **** | 0.5411 | **0.6669** | 0.000 | **** |
| **iliopsoas left** | 0.8189 | **0.8999** | 0.000 | **** | 0.7518 | **0.8637** | 0.000 | **** |
| **iliopsoas right** | 0.8053 | **0.8916** | 0.000 | **** | 0.7399 | **0.8403** | 0.000 | **** |
| **inferior vena cava** | 0.6087 | **0.7339** | 0.000 | **** | 0.4614 | **0.5991** | 0.000 | **** |
| **inner fat** | 0.5350 | **0.7684** | 0.000 | **** | 0.4366 | **0.6910** | 0.000 | **** |
| **intestine** | 0.6201 | **0.7818** | 0.000 | **** | 0.7069 | **0.7963** | 0.000 | **** |
| **IVD** | 0.6210 | **0.6615** | 0.000 | **** | 0.4118 | **0.5463** | 0.000 | **** |
| **kidney left** | 0.6541 | **0.8636** | 0.000 | **** | 0.5605 | **0.7452** | 0.000 | **** |
| **kidney right** | 0.6221 | **0.8753** | 0.000 | **** | 0.5938 | **0.6969** | 0.000 | **** |
| **liver** | 0.8242 | **0.9143** | 0.000 | **** | 0.7980 | **0.8722** | 0.000 | **** |
| **lung lower lobe left** | 0.7725 | **0.8343** | 0.000 | **** | 0.8056 | **0.9018** | 0.000 | **** |
| **lung lower lobe right** | 0.8127 | **0.8468** | 0.000 | **** | 0.8528 | **0.8858** | 0.000 | **** |
| **lung middle lobe right** | 0.7459 | **0.8313** | 0.000 | **** | 0.7381 | **0.8369** | 0.000 | **** |
| **lung upper lobe left** | 0.8706 | **0.9092** | 0.000 | **** | 0.8563 | **0.9171** | 0.000 | **** |
| **lung upper lobe right** | 0.8487 | **0.8924** | 0.000 | **** | 0.8538 | **0.9121** | 0.000 | **** |
| **muscle** | 0.8231 | **0.9647** | 0.000 | **** | 0.7978 | **0.9658** | 0.000 | **** |
| **pancreas** | 0.5243 | **0.6298** | 0.000 | **** | 0.3208 | **0.3483** | 0.000 | **** |
| **portal vein and splenic vein** | 0.3939 | **0.5564** | 0.000 | **** | 0.2833 | 0.2775 | 1.000 | ns |
| **prostate** | 0.6175 | **0.6759** | 0.000 | **** | NA | NA | NA | NA |
| **pulmonary vein** | 0.5851 | **0.6217** | 0.000 | **** | 0.5908 | **0.6068** | 0.000 | **** |
| **sacrum** | 0.8403 | **0.8963** | 0.000 | **** | 0.8040 | **0.8592** | 0.000 | **** |
| **scapula left** | 0.7722 | **0.9411** | 0.000 | **** | 0.7003 | **0.8947** | 0.000 | **** |
| **scapula right** | 0.7284 | **0.8781** | 0.000 | **** | 0.6921 | **0.9077** | 0.000 | **** |
| **spinal channel** | 0.6909 | **0.7500** | 0.000 | **** | 0.4940 | **0.7259** | 0.000 | **** |
| **spinal cord** | 0.5622 | **0.6762** | 0.000 | **** | 0.4678 | **0.6697** | 0.000 | **** |
| **spleen** | 0.4957 | **0.7064** | 0.000 | **** | 0.5162 | **0.6985** | 0.000 | **** |
| **sternum** | 0.7157 | 0.7286 | 1.000 | ns | 0.6461 | **0.8474** | 0.000 | **** |
| **stomach** | 0.5026 | **0.6361** | 0.000 | **** | 0.4130 | **0.5070** | 0.000 | **** |
| **subclavian artery left** | 0.5667 | **0.7504** | 0.000 | **** | 0.5318 | **0.6708** | 0.000 | **** |
| **subclavian artery right** | 0.5296 | **0.7990** | 0.000 | **** | 0.5403 | **0.7993** | 0.000 | **** |
| **subcutaneous fat** | 0.7751 | **0.9628** | 0.000 | **** | 0.8240 | **0.9756** | 0.000 | **** |
| **superior vena cava** | 0.6578 | **0.7281** | 0.000 | **** | 0.6016 | **0.7826** | 0.000 | **** |
| **thyroid gland** | 0.5087 | **0.8294** | 0.000 | **** | 0.4455 | **0.8086** | 0.000 | **** |
| **trachea** | 0.6730 | **0.7985** | 0.000 | **** | 0.5838 | **0.7455** | 0.000 | **** |
| **urinary bladder** | 0.5699 | **0.6433** | 0.000 | **** | 0.6191 | **0.7086** | 0.000 | **** |
| **vertebra body** | 0.7601 | **0.8001** | 0.000 | **** | 0.6032 | **0.7252** | 0.000 | **** |
| **vertebra posterior elements** | 0.6488 | 0.6495 | 1.000 | ns | 0.4961 | **0.6329** | 0.000 | **** |

Supplementary Table L: Mean dice scores of the 40 masks segmented with MRSegmentator for the female and male cohorts for the mask-supported and MIRTK registrations. The statistical tests were done using pairwise Wilcoxon signed-rank tests with adjustment for multiple testing using Bonferroni correction. (na: not applicable, ns: not significant, *: p-value < 0.05, **: p-value < 0.01, ***: p-value < 0.001, ****: p-value < 0.0001)

| | **male** | | | | **female** | | | |
|---|---|---|---|---|---|---|---|---|
| **mask name** | **MIRTK** | **mask supported** | **p.adj** | **significance** | **MIRTK** | **mask supported** | **p.adj** | **significance** |
| **aorta** | 0.6884 | **0.8476** | 0.0000 | **** | 0.6693 | **0.7701** | 0.0000 | **** |
| **colon** | 0.3158 | **0.5122** | 0.0000 | **** | **0.4404** | 0.4312 | 0.0000 | **** |
| **duodenum** | 0.3712 | **0.5616** | 0.0000 | **** | 0.3257 | **0.4316** | 0.0000 | **** |
| **esophagus** | 0.4342 | **0.6550** | 0.0000 | **** | 0.5194 | **0.5689** | 0.0000 | **** |
| **gallbladder** | 0.1513 | **0.2485** | 0.0000 | **** | **0.1671** | 0.1198 | 0.0000 | **** |
| **heart** | 0.7938 | **0.8880** | 0.0000 | **** | 0.8424 | **0.8835** | 0.0000 | **** |
| **inferior_vena_cava** | 0.5072 | **0.6760** | 0.0000 | **** | 0.3798 | **0.4965** | 0.0000 | **** |
| **left_adrenal_gland** | 0.3042 | **0.5170** | 0.0000 | **** | NA | NA | NA | NA |
| **left_autochthonous_muscle** | 0.8754 | **0.9102** | 0.0000 | **** | 0.8944 | **0.9017** | 0.0000 | **** |
| **left_femur** | 0.8541 | **0.8838** | 0.0000 | **** | 0.8775 | **0.9151** | 0.0000 | **** |
| **left_gluteus_maximus** | 0.8396 | **0.9234** | 0.0000 | **** | 0.9162 | **0.9247** | 0.0000 | **** |
| **left_gluteus_medius** | 0.8528 | **0.8878** | 0.0000 | **** | 0.8562 | **0.8910** | 0.0000 | **** |
| **left_gluteus_minimus** | 0.7790 | **0.8054** | 0.0000 | **** | 0.7315 | **0.7720** | 0.0000 | **** |
| **left_hip** | 0.8037 | **0.8676** | 0.0000 | **** | 0.7862 | **0.8706** | 0.0000 | **** |
| **left_iliac_artery** | 0.2369 | **0.4966** | 0.0000 | **** | 0.3305 | **0.4952** | 0.0000 | **** |
| **left_iliac_vena** | 0.3771 | **0.6530** | 0.0000 | **** | 0.5197 | **0.6830** | 0.0000 | **** |
| **left_iliopsoas_muscle** | 0.8112 | **0.8873** | 0.0000 | **** | 0.8144 | **0.8490** | 0.0000 | **** |
| **left_kidney** | 0.6588 | **0.8672** | 0.0000 | **** | 0.7224 | **0.7473** | 0.0000 | **** |
| **left_lung** | 0.8603 | **0.9097** | 0.0000 | **** | **0.9512** | 0.9390 | 0.0000 | **** |
| **liver** | 0.8292 | **0.9091** | 0.0000 | **** | 0.8533 | **0.8673** | 0.0000 | **** |
| **pancreas** | 0.4045 | **0.5279** | 0.0000 | **** | 0.2786 | **0.3001** | 0.0000 | **** |
| **portal_vein_and_splenic_vein** | 0.2334 | **0.3713** | 0.0000 | **** | **0.1907** | 0.1479 | 0.0000 | **** |
| **right_adrenal_gland** | 0.2794 | **0.4989** | 0.0000 | **** | 0.1366 | **0.2147** | 0.0000 | **** |
| **right_autochthonous_muscle** | 0.8599 | **0.8982** | 0.0000 | **** | **0.8927** | 0.8869 | 0.0000 | **** |
| **right_femur** | 0.8538 | **0.8798** | 0.0000 | **** | 0.8727 | **0.9110** | 0.0000 | **** |
| **right_gluteus_maximus** | 0.8802 | **0.9360** | 0.0000 | **** | 0.9084 | **0.9260** | 0.0000 | **** |
| **right_gluteus_medius** | 0.8664 | **0.9046** | 0.0000 | **** | 0.8523 | **0.8895** | 0.0000 | **** |
| **right_gluteus_minimus** | 0.7952 | **0.8237** | 0.0000 | **** | 0.7661 | **0.7881** | 0.0000 | **** |
| **right_hip** | 0.8112 | **0.8763** | 0.0000 | **** | 0.7684 | **0.8613** | 0.0000 | **** |
| **right_iliac_artery** | 0.2038 | **0.4013** | 0.0000 | **** | 0.2639 | **0.4856** | 0.0000 | **** |
| **right_iliac_vena** | 0.3269 | **0.5453** | 0.0000 | **** | 0.3789 | **0.5931** | 0.0000 | **** |
| **right_iliopsoas_muscle** | 0.7836 | **0.8819** | 0.0000 | **** | 0.8061 | **0.8372** | 0.0000 | **** |
| **right_kidney** | 0.6215 | **0.8824** | 0.0000 | **** | **0.7831** | 0.6877 | 0.0000 | **** |

| **right_lung** | 0.8699 | **0.9223** | 0.0000 | **** | **0.9624** | 0.9406 | 0.0000 | **** |
|---|---|---|---|---|---|---|---|---|
| **sacrum** | 0.7696 | **0.8482** | 0.0000 | **** | 0.7061 | **0.8101** | 0.0000 | **** |
| **small_bowel** | 0.3919 | **0.5639** | 0.0000 | **** | 0.5015 | **0.5325** | 0.0000 | **** |
| **spine** | 0.7528 | **0.7831** | 0.0000 | **** | 0.6142 | **0.6946** | 0.0000 | **** |
| **spleen** | 0.5041 | **0.7186** | 0.0000 | **** | 0.5924 | **0.7002** | 0.0000 | **** |
| **stomach** | 0.4603 | **0.5982** | 0.0000 | **** | 0.4224 | **0.4616** | 0.0000 | **** |
| **urinary_bladder** | 0.5799 | **0.6291** | 0.0000 | **** | **0.8065** | 0.6687 | 0.0000 | **** |

Supplementary Table M: Mean dice scores of the 50 masks segmented with TotalSegmentator for the female and male cohorts for the mask-supported and MIRTK registrations. The statistical tests were done using pairwise Wilcoxon signed-rank tests with adjustment for multiple testing using Bonferroni correction. (na: not applicable, ns: not significant, *: p-value < 0.05, **: p-value < 0.01, ***: p-value < 0.001, ****: p-value < 0.0001)

| | male | | | | female | | | |
|---|---|---|---|---|---|---|---|---|
| mask name | MIRTK | mask supported | p.adj | significance | MIRTK | mask supported | p.adj | significance |
| adrenal_gland_left | 0.2696 | **0.4339** | 0.0000 | **** | 0.0623 | **0.1384** | 0.0000 | **** |
| adrenal_gland_right | 0.2518 | **0.4516** | 0.0000 | **** | 0.0853 | **0.2034** | 0.0000 | **** |
| aorta | 0.6183 | **0.8005** | 0.0000 | **** | 0.5296 | **0.7062** | 0.0000 | **** |
| autochthon_left | 0.8710 | **0.9068** | 0.0000 | **** | 0.8146 | **0.8929** | 0.0000 | **** |
| autochthon_right | 0.8526 | **0.8994** | 0.0000 | **** | 0.7914 | **0.8762** | 0.0000 | **** |
| clavicula_left | 0.4271 | **0.5717** | 0.0000 | **** | 0.3589 | **0.5150** | 0.0000 | **** |
| clavicula_right | 0.3905 | **0.5050** | 0.0000 | **** | 0.3447 | **0.5156** | 0.0000 | **** |
| colon | 0.2882 | **0.4759** | 0.0000 | **** | 0.2619 | **0.3747** | 0.0000 | **** |
| duodenum | 0.2870 | **0.4763** | 0.0000 | **** | 0.1481 | **0.2574** | 0.0000 | **** |
| esophagus | 0.4220 | **0.6156** | 0.0000 | **** | 0.3999 | **0.5442** | 0.0000 | **** |
| femur_left | 0.8380 | **0.8722** | 0.0000 | **** | 0.8256 | **0.8972** | 0.0000 | **** |
| femur_right | 0.8414 | **0.8758** | 0.0000 | **** | 0.8370 | **0.9038** | 0.0000 | **** |
| gallbladder | 0.1395 | **0.2269** | 0.0000 | **** | 0.0611 | **0.1199** | 0.0000 | **** |
| gluteus_maximus_left | 0.8415 | **0.9145** | 0.0000 | **** | 0.8510 | **0.9189** | 0.0000 | **** |
| gluteus_maximus_right | 0.8686 | **0.9178** | 0.0000 | **** | 0.7817 | **0.9069** | 0.0000 | **** |
| gluteus_medius_left | 0.7958 | **0.8269** | 0.0000 | **** | 0.6907 | **0.7589** | 0.0000 | **** |
| gluteus_medius_right | 0.8077 | **0.8455** | 0.0000 | **** | 0.6390 | **0.7458** | 0.0000 | **** |
| gluteus_minimus_left | 0.7391 | **0.7703** | 0.0000 | **** | 0.6550 | **0.7179** | 0.0000 | **** |
| gluteus_minimus_right | 0.7352 | **0.7644** | 0.0000 | **** | 0.6436 | **0.7219** | 0.0000 | **** |
| heart | 0.7826 | **0.8678** | 0.0000 | **** | 0.7900 | **0.8632** | 0.0000 | **** |
| hip_left | 0.7274 | **0.8251** | 0.0000 | **** | 0.6486 | **0.8027** | 0.0000 | **** |
| hip_right | 0.7362 | **0.8312** | 0.0000 | **** | 0.5901 | **0.7971** | 0.0000 | **** |
| humerus_left | 0.7410 | **0.8263** | 0.0000 | **** | 0.6942 | **0.8010** | 0.0000 | **** |
| humerus_right | **0.7116** | 0.6805 | 0.0000 | **** | 0.6359 | **0.8181** | 0.0000 | **** |
| iliac_artery_left | 0.2005 | **0.4032** | 0.0000 | **** | 0.1809 | **0.3290** | 0.0000 | **** |
| iliac_artery_right | 0.1702 | **0.3359** | 0.0000 | **** | 0.1805 | **0.3546** | 0.0000 | **** |
| iliac_vena_left | 0.3268 | **0.5562** | 0.0000 | **** | 0.3542 | **0.5605** | 0.0000 | **** |
| iliac_vena_right | 0.2882 | **0.5024** | 0.0000 | **** | 0.3446 | **0.4995** | 0.0000 | **** |
| iliopsoas_left | 0.7681 | **0.8563** | 0.0000 | **** | 0.6931 | **0.8243** | 0.0000 | **** |
| iliopsoas_right | 0.7523 | **0.8666** | 0.0000 | **** | 0.6756 | **0.8114** | 0.0000 | **** |
| inferior_vena_cava | 0.4507 | **0.5723** | 0.0000 | **** | 0.1336 | **0.2391** | 0.0000 | **** |
| intervertebral_discs | 0.5953 | **0.6174** | 0.0000 | **** | 0.3902 | **0.5059** | 0.0000 | **** |
| kidney_left | 0.5677 | **0.8224** | 0.0000 | **** | 0.4288 | **0.6746** | 0.0000 | **** |

| **kidney_right** | 0.5443 | **0.8291** | 0.0000 | **** | 0.4806 | **0.6057** | 0.0000 | **** |
|---|---|---|---|---|---|---|---|---|
| **liver** | 0.8101 | **0.8918** | 0.0000 | **** | 0.7915 | **0.8498** | 0.0000 | **** |
| **lung_left** | 0.8538 | **0.9086** | 0.0000 | **** | 0.8590 | **0.9362** | 0.0000 | **** |
| **lung_right** | 0.8643 | **0.9229** | 0.0000 | **** | 0.8866 | **0.9445** | 0.0000 | **** |
| **pancreas** | 0.2529 | **0.3653** | 0.0000 | **** | 0.0476 | **0.0588** | 0.0000 | **** |
| **portal_vein_and_splenic_vein** | 0.1735 | **0.2966** | 0.0000 | **** | **0.0783** | 0.0704 | 0.0480 | * |
| **prostate** | 0.5628 | **0.6333** | 0.0000 | **** | NA | NA | NA | NA |
| **sacrum** | 0.6799 | **0.7583** | 0.0000 | **** | 0.6370 | **0.7094** | 0.0000 | **** |
| **scapula_left** | 0.5832 | **0.7053** | 0.0000 | **** | 0.5117 | **0.6530** | 0.0000 | **** |
| **scapula_right** | 0.5384 | **0.6134** | 0.0000 | **** | 0.5213 | **0.6597** | 0.0000 | **** |
| **small_bowel** | 0.2511 | **0.3959** | 0.0000 | **** | 0.2344 | **0.2944** | 0.0000 | **** |
| **spinal_cord** | 0.7145 | **0.7679** | 0.0000 | **** | 0.4775 | **0.7477** | 0.0000 | **** |
| **spleen** | 0.4866 | **0.6994** | 0.0000 | **** | 0.4746 | **0.6618** | 0.0000 | **** |
| **stomach** | 0.4375 | **0.5762** | 0.0000 | **** | 0.3404 | **0.4374** | 0.0000 | **** |
| **urinary_bladder** | 0.5808 | **0.6373** | 0.0000 | **** | 0.5824 | **0.6485** | 0.0000 | **** |
| **vertebrae** | 0.7143 | **0.7413** | 0.0000 | **** | 0.5444 | **0.6640** | 0.0000 | **** |